\documentclass{article}

\usepackage{booktabs}
\usepackage{amsthm}
\usepackage{caption}
\usepackage{subcaption}

\usepackage[preprint,nonatbib]{neurips_2022}
\newtheorem{theorem}{Theorem}[section]
\newtheorem{lemma}{Lemma}[section]

\usepackage[utf8]{inputenc} 
\usepackage[T1]{fontenc}    
\usepackage{hyperref}       
\usepackage{url}            
\usepackage{booktabs}       
\usepackage{amsfonts}       
\usepackage{nicefrac}       
\usepackage{microtype}      
\usepackage[dvipsnames]{xcolor}         
\usepackage{graphicx}

\usepackage{amsmath}
\usepackage{amssymb}

\def\*#1{\mathbf{#1}}
\newcommand{\re}[1]{\ensuremath{\mathrm{Re}\left(#1\right)}}
\newcommand{\im}[1]{\ensuremath{\mathrm{Im}\left(#1\right)}}

\usepackage{lipsum}
\newcommand\blfootnote[1]{%
  \begingroup
  \renewcommand\thefootnote{}\footnote{#1}%
  \addtocounter{footnote}{-1}%
  \endgroup
}

\title{
Chefs' Random Tables: Non-Trigonometric Random Features}

\author{%
  Valerii Likhosherstov* \\
  University of Cambridge\\
  \texttt{vl304@cam.ac.uk} \\
  \And
  Krzysztof Choromanski* \\
  Google Research \& Columbia University\\
  \texttt{kchoro@google.com} \\
  \And
   Avinava Dubey*\\
  Google Research \\
  \And
  Frederick Liu*\\
  Google Research \\
  \And
  Tamas Sarlos\\
  Google Research \\
  \And
  Adrian Weller\\
  University of Cambridge \& \\ The Alan Turing Institute\\
}

\begin{document}

\maketitle

\begin{abstract}
We introduce \textit{chefs' random tables} (CRTs), a new class of non-trigonometric random features (RFs) to approximate Gaussian and softmax-kernels. CRTs are an alternative to standard random kitchen sink (RKS) methods, which inherently rely on the trigonometric maps \cite{rfs3}. 
We present variants of CRTs where RFs are positive, a key requirement for applications in recent low-rank Transformers \cite{performer}. Further variance reduction is possible by leveraging statistics which are simple to compute. One instantiation of CRTs, the \textit{optimal positive random features} (OPRFs), is to our knowledge the first RF method for unbiased softmax-kernel estimation with positive and bounded RFs, resulting in exponentially small tails and 
much lower variance 
than its counterparts. As we show,
orthogonal random features applied in OPRFs provide additional variance reduction for any dimensionality $d$ (not only asymptotically for sufficiently large $d$, as for RKS). We 
test CRTs on many tasks ranging from non-parametric classification to training Transformers for text, speech and image data, obtaining new state-of-the-art results for low-rank text Transformers, while providing linear space and time complexity of the attention.
\end{abstract}

\section{Introduction \& related work}

\blfootnote{* Equal contribution}The idea that nonlinear mappings of the
random-weight linear combinations of data features can be used to linearize various nonlinear similarity functions transformed kernel methods. This led to the development of 
\textit{Random Kitchen Sinks} (RKSs) techniques; and the new field of scalable kernel algorithms, introduced in the paper trilogy \cite{rfs2, rfs, rfs3}, was born. RKSs were subsequently used in many applications, ranging from kernel and function-to-function regression \cite{kernel-ridge-rfs, laparra, oliva}, SVM algorithms \cite{svmrfs} to operator-valued and semigroup kernels \cite{minh, yang},  neural networks \cite{nnrfs, xie, cho, han} and even differentially-private ML algorithms \cite{sarwate}, as well as (very recently) nonparametric adaptive control \cite{boffi}. Random features (RFs) 
are a subject of 
much theoretical analysis \cite{liton, nystromrfs, sutherland, szabo}.

To approximate shift invariant (e.g. Gaussian, Cauchy or Laplace) and softmax kernels, RKSs rely on the trigonometric nonlinear mappings provided directly by Bochner's Theorem \cite{minh}. Trigonometric RFs provide strong concentration results (e.g. uniform convergence, see Claim 1 in \cite{rfs}), but suffer from a weakness that 
was noted recently -- they are not guaranteed to be positive. This makes them unsuitable for approximating softmax-attention in scalable Transformers relying on implicit attention via random features \cite{performer}. As noted in \cite{performer}, trigonometric features lead to unstable training, as 
they yield poor approximations of the partition functions applied to renormalize attention and involving several small softmax kernel values. To address this, \cite{performer} proposed a new method for unbiased softmax kernel estimation with positive RFs, the so-called \textit{FAVOR+} mechanism (\textbf{F}ast \textbf{A}ttention \textbf{V}ia \textbf{O}rthogonal \textbf{R}andom Positive Features), as opposed to FAVOR using trigonometric RFs (as in  \cite{perf0}). 

\begin{figure}
     \centering
     \begin{subfigure}[b]{0.4\textwidth}
         \centering
         \includegraphics[width=\textwidth]{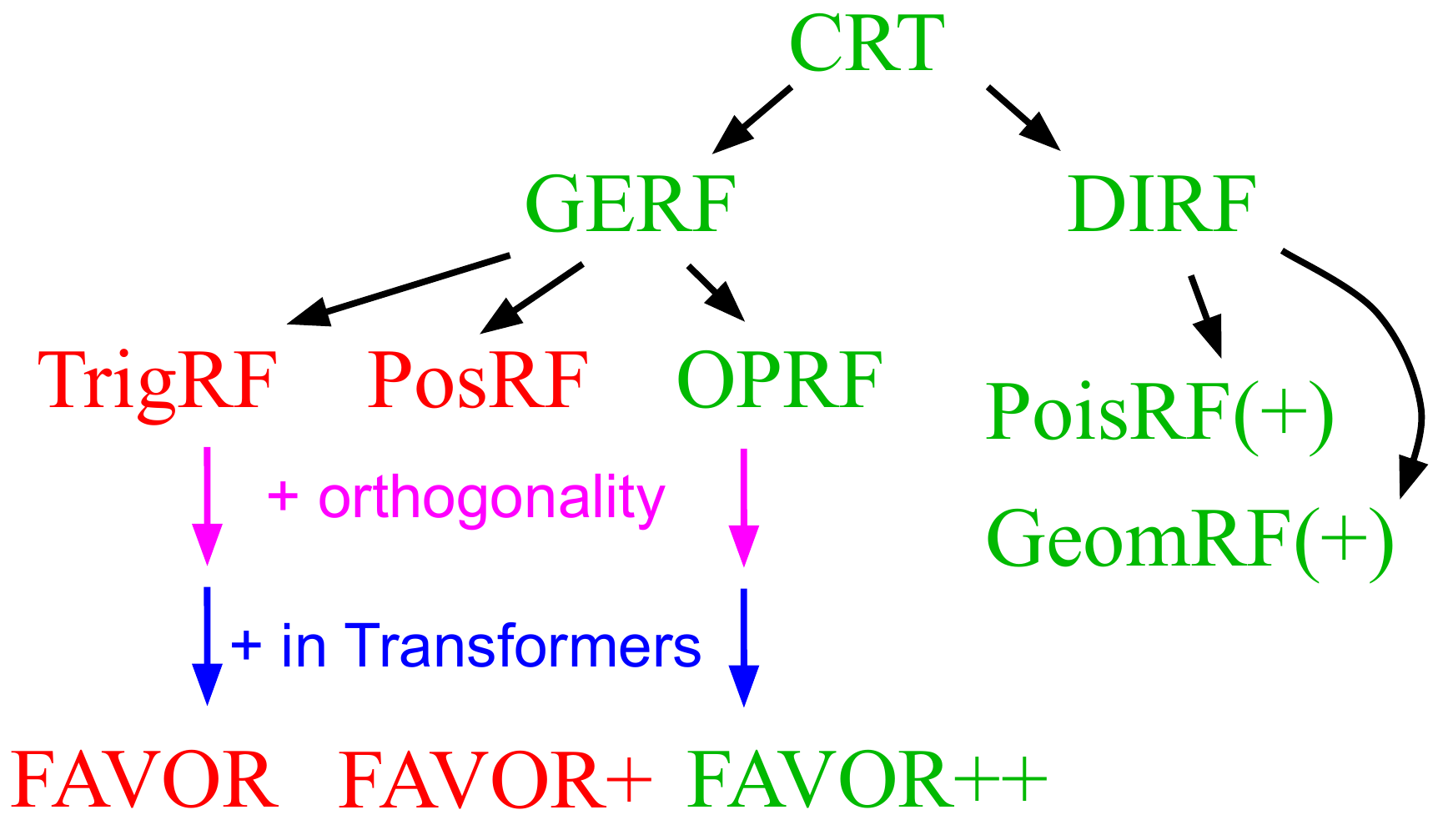}
     \end{subfigure}
     \hfill
     \begin{subfigure}[b]{0.59\textwidth}
         \centering
         \includegraphics[width=\textwidth]{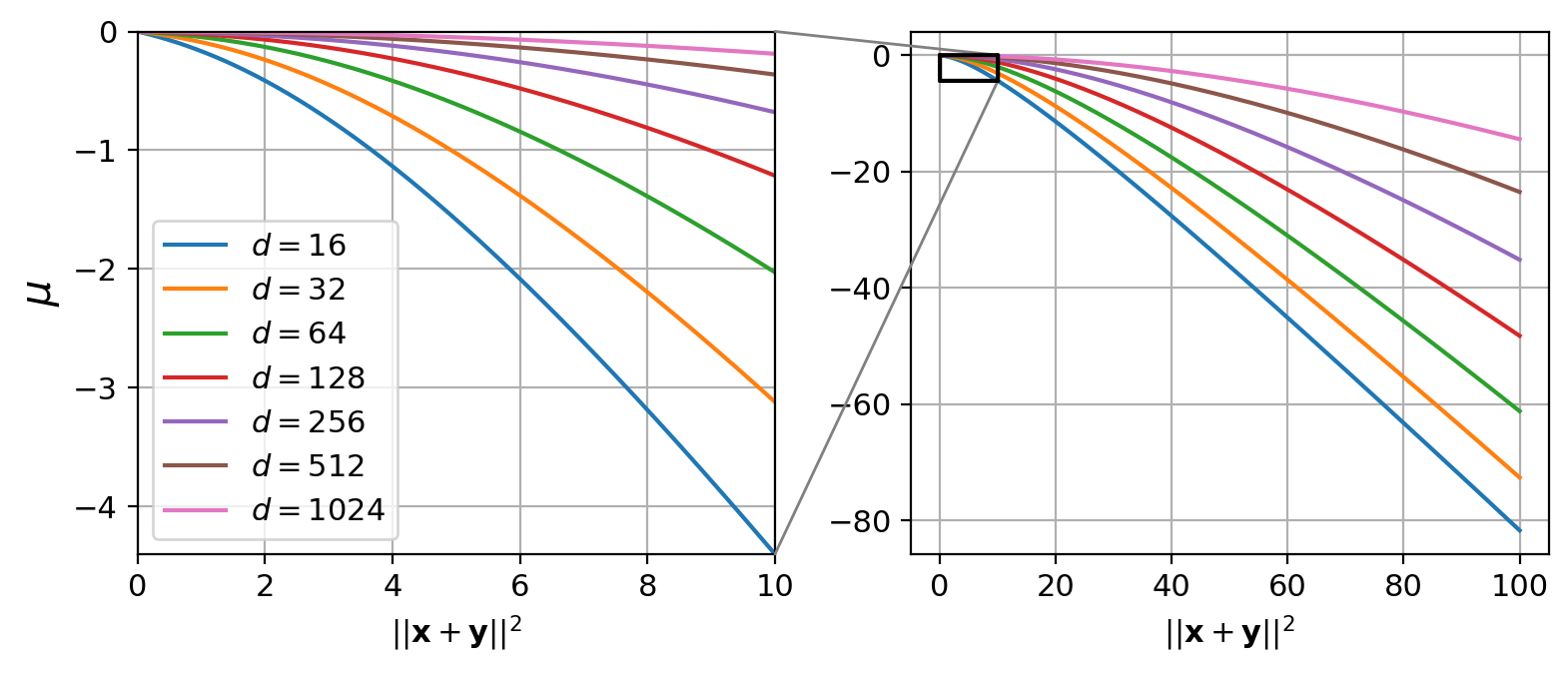}
     \end{subfigure}
        \caption{\small{\textbf{(left)} A map of RF methods for the Gaussian kernel approximation. {\color{red} Existing RFs (Section \ref{sec:exrfs})}, {\color{ForestGreen} RFs proposed in this paper}. \textbf{(right)} The utility function $\mu$ (defined as the logarithm of the ratio of the variance of OPRF and PosRF 
        mechanisms for the Gaussian and softmax kernel estimation) as a function of squared length of the sum of kernels' inputs $\|\mathbf{x}+\mathbf{y}\|^{2}$ (smaller values imply larger gains coming from OPRF). Different curves correspond to different dimensionalities. Based on the plots, OPRFs have $> e^{60}$ times smaller variance when $d = 64, \| \*x + \*y \|^2 = 100$ (configuration taken from the standard Transformer application).}} \vspace{-5mm}
        \label{fig:varratio}
\end{figure}



Unfortunately FAVOR+ features are not bounded, and worse, the moment generating function of the corresponding softmax kernel estimator is not finite. Consequently, no concentration results beyond those involving second moment methods (variance bounds) have been provided for FAVOR+. Despite  active research on constructing new RFs for implicit attention in Transformers \cite{hrfs, tr-kernel}, the following questions of great practical importance remained open:

\textit{Does there exist an unbiased estimator of the softmax/Gaussian kernel relying on positive and simultaneously bounded random features? Can it be efficiently constructed?} 

We  answer both questions affirmatively in this paper, introducing a new mechanism called \textit{optimal positive random features} (OPRFs). We propose other RF methods that, as OPRFs, do not apply trigonometric functions and provide positivity. We call this new set of RF mechanisms \textit{chefs' random tables} (CRTs, see Figure \ref{fig:varratio}-left). The new OPRF-based method for fast self-attention approximation, applying in addition block-orthogonal random projections, is referred to as \textit{FAVOR++}.

We compute the variance of OPRF-based estimators (see Theorem \ref{lemma:gerfs} \& \ref{lemma:gevar}) and show that they can provide $e^{60}$x variance reduction for Gaussian/softmax kernel estimation (see Figure \ref{fig:varratio}). We give the first exponentially small upper bounds for tails of the Gaussian/softmax kernel estimators relying on positive RFs, leveraging boundedness of OPRFs (see Theorem \ref{thm:tails}). Consequently, using OPRFs we give the first uniform convergence results for softmax attention approximation with positive RFs (Theorem \ref{thm:uniform}). We show that orthogonal random projections combined with OPRFs (leading to FAVOR++) provably reduce the variance of OPRFs for any dimensionality $d$ (see  Theorem \ref{thm-orthoprfs}) as opposed to only asymptotically for $d$ large enough which is the case for RKSs. Finally, we provide extensive empirical evaluation in Section \ref{sec:exps}, for Transformers (text, image and speech domains), establishing new state-of-the-art results for low-rank attention Transformers for text.
\section{Prerequisites}
\label{sec:prerequisites}

\subsection{The definition of random features}

Let $\*x, \*y \in \mathbb{R}^d$ be real vectors and $K(\*x, \*y) = \exp(\frac12 \| \*x - \*y \|^2)$ be a Gaussian kernel where $\|\cdot\|$ 
denotes the $L_2$-norm. 
By \textit{random features (RFs) for the Gaussian kernel} we denote two functions $f^{(1)}(\omega, \*x), f^{(2)}(\omega, \*y): \mathbb{R}^d \times \mathbb{R}^d \to \mathbb{C}$ where $\omega$ is a random vector from some distribution $p (\omega)$ on $\mathbb{R}^d$. Functions $f^{(\cdot)}(\omega, \*x)$ satisfy the following:
\begin{gather}
    K (\*x, \*y) = \mathbb{E}_{p(\omega)} \re{  f^{(1)}(\omega, \*x) f^{(2)}(\omega, \*y)} \label{eq:rf}
\end{gather}
for all $\*x, \*y \in \mathbb{R}^d$ where $\re{\cdot}$ denote the real part of a complex number ($\im{\cdot}$ for the imaginary part). The decomposition (\ref{eq:rf}) can be used for an unbiased approximation of the linear operator $\mathcal{K} = (K (\*x^{(i)}, \*y^{(j)}))_{i,j = 1}^{L,L} \in \mathbb{R}^{L \times L}$ where $\*x^{(i)}, \*y^{(j)} \in \mathbb{R}^d$. Such linear operators emerge in various applications, e.g. kernel SVM \cite{rfs}, kernel regression \cite{nadaraya,watson} or Transformers \cite{performer} (see Section \ref{sec:transf}).

For any $\*c \in \mathbb{R}^L$, evaluating $\mathcal{K} \*c$ naively would result in $O (d L^2)$ time complexity which is prohibitively expensive for large $L$. Instead, we can use the Monte Carlo approximation: draw i.i.d. samples $\omega_1, \dots, \omega_M \sim p(\omega)$, where $M \ll L$, and compute for $1 \leq i \leq L$:
\begin{gather}
    (\mathcal{K} \*c)_i = \sum_{j = 1}^L K(\*x^{(i)}, \*y^{(j)}) \*c_j \approx \sum_{j = 1}^L \left( \frac{1}{M} \sum_{m = 1}^M \re{ f^{(1)}(\omega_m, \*x^{(i)}) f^{(2)}(\omega_m, \*y^{(j)})} \right) \*c_j \nonumber \\
    = \frac{1}{M} \re{ \sum_{m = 1}^M f^{(1)}(\omega_m, \*x^{(i)}) \sum_{j = 1}^L f^{(2)}(\omega_m, \*y^{(j)}) \*c_j } . \label{eq:rfappr}
\end{gather}
Therefore, $\mathcal{K} \*c$ can be approximated by first precomputing $\{ \sum_{j = 1}^L f^{(2)}(\omega_m, \*y^{(j)})
\mathbf{c}_{j}\}_{m = 1}^M$ and then evaluating (\ref{eq:rfappr}) in $O (d M L)$ total time. Precision of this approximation can be theoretically bounded \cite{rfs,performer}. In this manuscript, we will use the variance $\mathrm{Var}_{p (\omega)} \re{  f^{(1)}(\omega, \*x) f^{(2)}(\omega, \*y)}$ of (\ref{eq:rf}) to judge the precision of the Monte Carlo approximation. Since samples $\omega_1, \dots, \omega_M$ are i.i.d., the number $M$ of RFs controls the tradeoff between the total variance $\mathrm{Var}_{\omega_1, \dots, \omega_M} (\dots) = \frac{1}{M} \mathrm{Var}_{p(\omega)} (\dots)$ 
(inversely proportional to $M$) and number of computations (directly proportional to $M$).

The softmax kernel is defined as: $K_{\mathrm{sfm}}(\mathbf{x},\mathbf{y})=\exp(\mathbf{x}^{T}\mathbf{y})$, and can be easily derived from the Gaussian kernel $K$ as follows: $K_{\mathrm{sfm}}(\mathbf{x},\mathbf{y}) = \exp(\|\mathbf{x}\|^{2} / 2)K(\mathbf{x},\mathbf{y})\exp(\|\mathbf{y}\|^{2} / 2)$. Thus, in particular, any RF mechanism for the Gaussian kernel immediately transfers to the corresponding one for the softmax kernel and vice versa. Thus from now on, unless explicitly stated otherwise, 
the estimators we consider are approximating the Gaussian kernel. 

\subsection{Existing trigonometric and positive random feature methods} \label{sec:exrfs}

Here we summarize existing RFs for Gaussian kernel estimation. \textit{Trigonometric RFs (TrigRFs)}, the core of RKSs \cite{rfs} and FAVOR \cite{perf0}, are defined as follows: $f^{(1)}_\mathrm{trig} (\omega, \*x) = \exp (\mathrm{i} \omega^\top \*x)$, $f^{(2)}_\mathrm{trig} (\omega, \*y) = \exp (- \mathrm{i} \omega^\top \*y)$, $p_\mathrm{trig} (\omega) \sim \mathcal{N} (\*0_d, \*I_d)$ where $\mathrm{i}$ denotes an imaginary unit (as opposed to the index notation $i$), $\*0_d \in \mathbb{R}^d$ is a vector of zeros and $\*I_d \in \mathbb{R}^{d \times d}$ is an identity matrix. The variance of these RFs has the following form \cite{performer}: $\mathrm{Var}_{p_\mathrm{trig} (\omega)} \re{f^{(1)}_\mathrm{trig} (\omega, \*x) f^{(2)}_\mathrm{trig} (\omega, \*y)} = \frac12 \left( 1 - K (\*x, \*y)^2 \right)^2$.

\textit{Positive RFs (PosRFs)} \cite{performer}, the key ingredient of the FAVOR+ mechanism, are defined as follows: $f^{(1)}_\mathrm{pos} (\omega, \*x) = f^{(2)}_\mathrm{pos} (\omega, \*x) = \exp (\omega^\top \*x - \| \*x \|^2)$, $p_\mathrm{pos} (\omega) \sim \mathcal{N} (\*0_d, \*I_d)$. Their name is due to the fact that $f^{(1)}_\mathrm{pos} (\omega, \*x), f^{(2)}_\mathrm{pos} (\omega, \*y)$ are always positive real numbers. PosRF variance has the form \cite{performer}: $\mathrm{Var}_{p_\mathrm{pos} (\omega)} \left( f^{(1)}_\mathrm{pos} (\omega, \*x) f^{(2)}_\mathrm{pos} (\omega, \*y) \right) = \exp (4 \*x^\top \*y) - K (\*x, \*y)^2$.

\subsection{Random features for scalable Transformers} \label{sec:transf}

One recent application of RFs is in the area of scalable Transformers for processing long sequences \cite{performer}. Let $L$ be the length of the sequence. Interactions between elements in Transformers are implemented via the \textit{self-attention mechanism}. Given three matrices $\*Q, \*K, \*V \in \mathbb{R}^{L \times d}$, the self-attention mechanism returns the following result:
\begin{equation}
    \*Y = \mathrm{softmax} (d^{- 1 / 2} \*Q \*K^\top) \*V = \mathrm{diag}(\mathcal{K}_\mathrm{sfm} \mathbf{1}_L)^{-1} \mathcal{K}_\mathrm{sfm} \*V, \quad \mathcal{K}_\mathrm{sfm} =  (K_\mathrm{sfm} (\*x_i, \*y_j ) )_{i,j = 1}^{L, L} \label{eq:sadef} 
\end{equation}
where $\mathbf{1}_L \in \mathbb{R}^L$ is a vector of all ones, 
$\*x^{(i)} = d^{- 1 / 4} \*Q_{i,:}$ ($i$'th row of $\*Q$) and $\*y^{(j)} = d^{- 1 / 4} \*K_{j,:}$, $1 \leq i,j \leq L$.
We deduce that 
computing (\ref{eq:sadef}) reduces to applying the linear operator $\mathcal{K}_\mathrm{sfm}$ to $d + 1$ vectors: $\*1_L$, $\*V_{:,1}$, \dots, $\*V_{:,d}$. Hence, when $L$ is large, RF approximation similar to (\ref{eq:rfappr}) but for the $K_\mathrm{sfm} (\cdot, \cdot)$ kernel can reduce the computational complexity from $O(d L^2)$ to $O (d M L)$.


Importantly, when the approximation (\ref{eq:rfappr}) with the replacement $\mathcal{K} \to \mathcal{K}_\mathrm{sfm}$ 
can take negative and/or near-zero values, training is unstable since this approximation 
emerges in the denominator (inversed) term $\mathrm{diag} (\mathcal{K}_\mathrm{sfm} \*1_L)$ in (\ref{eq:sadef}). One way to address this is to restrict $f^{(1)} (\omega, \*x)$ and $f^{(2)} (\omega, \*y)$ to always map into strictly positive numbers $\mathbb{R}^+$. This is where PosRFs introduced in \ref{sec:exrfs} are particularly relevant.

\section{Chefs' Random Tables}
\label{sec:crt}

We are ready to present our mechanism of chefs' random tables. {All proofs are in the Appendix.}

\subsection{Generalized exponential RFs (GERFs) \& optimal positive RFs (OPRFs)} 
\label{sec:gexp}

Our first goal will be to generalize both trigonometric and positive RFs. Then we will focus on one special case of this generalization, that will directly lead to the FAVOR++ mechanism.

We will be looking for RFs of the following generalized exponential form for $p_\mathrm{GE} (\omega) \sim \mathcal{N} (\*0_d, \*I_d)$:
\begin{align}
\begin{split}
\label{eq:f-func}
f^{(1)}_\mathrm{GE} (\omega, \mathbf{x}) = D \exp( A \| \omega \|^2 + B \omega^\top \*x + C \| \*x \|^2 ), \\
f^{(2)}_\mathrm{GE} (\omega, \*y) = D \exp (A \| \omega \|^2 + s B \omega^\top \*y + C \| \*y \|^2),
\end{split}
\end{align}
where $A, B, C, D \in \mathbb{C}$ and $s \in \{ -1, +1 \}$. It can be seen that $A = 0$, $B = \mathrm{i}$, $C = 0$, $D = 1$, $s = -1$ corresponds to trigonometric RFs and $A = 0$, $B = 1$, $C = -1$, $D = 1$, $s = 1$ corresponds to positive RFs. The next theorem describes the conditions under which $f^{(.)}_\mathrm{GE}$ can be used to approximate the Gaussian kernel. 

\begin{theorem} \label{lemma:gerfs}
$p_\mathrm{GE} (\omega)$ and $f^{(.)}_\mathrm{GE}$, defined in (\ref{eq:f-func}), satisfy (\ref{eq:rf}) if
\begin{equation}
    \re{1 - 4 A} > 0, \quad B = \sqrt{s (1 - 4 A)}, \quad C = - (s + 1) / 2, \quad D = (\sqrt[4]{1 - 4 A})^d, \label{eq:abcd}
\end{equation}
where $\sqrt{\cdot}$ and $\sqrt[n]{\cdot}$ denotes a \textit{principal root} if the argument is complex.
\end{theorem}

Hence, $A$ and $s$ can be treated as free parameters and $B$, $C$, $D$ as dependent ones. The variance of these RFs can be expressed through $A$ and $s$ as follows:
\begin{theorem} \label{lemma:gevar}
Let $\re{1 - 8 A} > 0$.
The variance of \eqref{eq:rfappr} using $p_\mathrm{GE} (\omega), f^{(.)}_\mathrm{GE}$ is given as
\begin{align}
    &\mathrm{Var}_{p_\mathrm{GE} (\omega)} \re{f^{(1)}_\mathrm{GE} (\omega, \*x) f^{(2)}_\mathrm{GE} (\omega, \*y)} = \frac12 \exp \left( - (s + 1) \left( \| \*x \|^2 + \| \*y \|^2 \right) \right) \nonumber \\
    &\times \biggl( \mathrm{Re} \biggl( \alpha_1 \exp \biggl( \alpha_2 \| \*x + s \*y \|^2 \biggr) \biggr) + \alpha_3 \exp \biggl( \alpha_4 \| \*x + s \*y \|^2 \biggr) \biggr) - K (\*x, \*y)^2 . \label{eq:gexpvar}
\end{align}

where {\small{$\alpha_1 = \left(\sqrt{1 + \frac{16 A^2}{1 - 8 A}} \right)^d$, $\alpha_2 = \left( s + \frac{s}{1 - 8 A} \right)$, $\alpha_3 = \left( 1 + \frac{16 | A |^2}{1 - 8 \re{A}} \right)^{d / 2}$, $\alpha_4  = \left( \frac{s}{2} + \frac{s + 2 | 1 - 4 A |}{2( 1 - 8 \re{A})} \right)$.} }
\end{theorem}


While it is  unclear how to find a global minimum of the objective (\ref{eq:gexpvar}) with respect to $A \in \mathbb{C}$, $\re{1 - 8 A} > 0$ and $s \in \{ -1, +1 \}$, we observe that it's possible to find an optimum when we restrict $A$ to be a real number and fix $s = +1$. 

\begin{theorem}[Minimum variance] \label{lemma:minvar}
When $s=+1$, $A$ is restricted to be a real number and $\| \*x + \*y \|^2 > 0$, the variance \eqref{eq:gexpvar} is minimized when $A = (1 - 1 / \rho^*) / 8$ where
\begin{equation}
    \rho^* = \left( \sqrt{\left( 2 \| \*x + \*y \|^2 + d \right)^2 + 8 d\| \*x + \*y \|^2} - 2 \| \*x + \*y \|^2 - d\right) / \left( 4 \| \*x + \*y \|^2 \right). \label{eq:rho1}
\end{equation}
\end{theorem}
\textbf{Note:} One can show that $A < 0$ for $\mathbf{x} \neq -\mathbf{y}$ thus the corresponding estimator is bounded since the term $A \| \omega \|^2$ prevails over linear terms $B \omega^\top \*x$ and $sB \omega^\top \*y$ in (\ref{eq:f-func}). When $\|\mathbf{x}+\mathbf{y}\| \rightarrow 0$ then $\rho^{*} \rightarrow 1$ and thus $A \rightarrow 0$. Therefore for $\mathbf{x}=-\mathbf{y}$ the mechanism reduces to PosRF described in Sec. \ref{sec:exrfs} as expected, since for $\mathbf{x}=-\mathbf{y}$ PosRFs provide perfect estimation (variance equal to zero). Larger values of $\|\mathbf{x}+\mathbf{y}\|$ lead to larger gains coming from the new mechanism.

From (\ref{eq:abcd}) it can be inferred that $B, C, D$ are real when $A$ is real and $s = +1$. Hence, $f^{(1)} (\omega, \*x)$, $f^{(2)}(\omega, \*y)$ are positive real numbers in this case. Furthermore, $s = +1, A = 0$ corresponds to positive RFs. Therefore, we refer to RFs with $A$ defined according to (\ref{eq:rho1}) as \textit{optimal positive RFs} (OPRFs). Figure \ref{fig:varratio}-right illustrates the analytical variance reduction achieved via OPRFs. 

In practice, we are given sets $\{ \*x^{(i)} \}$, $\{ \*y^{(j)} \}$ instead of a single pair $\*x, \*y$. For this reason, in (\ref{eq:gexpvar},\ref{eq:rho1}), we can use the averages of $\| \*x^{(i)} \|^2$, $\| \*y^{(j)} \|^2$, $\| \*x^{(i)} + s \*y^{(j)} \|^2$ instead of $\| \*x \|^2$, $\| \*y \|^2$, $\| \*x + s \*y \|^2$. This heuristic is based on the assumption that all $\{ \*x^{(i)} \}$ and $\{ \*y^{(j)} \}$ are homogeneous and $\| \*x^{(i)} \|^2$, $\| \*y^{(j)} \|^2$, $\| \*x^{(i)} + s \*y^{(j)} \|^2$ are tightly concentrated around their mean. Computing averages of $\| \*x^{(i)} \|^2$, $\| \*y^{(j)} \|^2$ takes $O(L d)$ time. Using the formula below, the average of $\| \*x^{(i)} + s \*y^{(j)} \|^2$ can be computed with the same complexity:
\begin{equation}
    \frac{1}{L^2} \sum_{i = 1}^L \sum_{j = 1}^L \| \*x_i + s \*y_j \|^2 = \frac{1}{L} \sum_{i = 1}^L \| \*x_i \|^2 + \frac{2s}{L^2} \left( \sum_{i = 1}^L \*x_i \right)^\top \left( \sum_{i = 1}^L \*y_i \right) + \frac{1}{L} \sum_{i = 1}^L \| \*y_i \|^2 . \label{eq:gestat}
\end{equation}
The closed-form solution for real $A$ and $s = +1$ allows $O(1)$-time optimization of (\ref{eq:gexpvar}) after precomputing these statistics. In the general case we can rely on numerical optimization of (\ref{eq:gexpvar}) with respect to $A \in \mathbb{C}$ and $s \in \{ -1, +1 \}$. Using precomputed statistics, each evaluation of (\ref{eq:gexpvar}) takes $O(1)$ time. As long as the total number of these evaluations is $O (L M (d + n))$, where $n$ is the number of $\mathcal{K}$ or $\mathcal{K}_\mathrm{sfm}$ evaluations ($n = d + 1$ in Section \ref{sec:transf}), it does not affect the total complexity. 


The next 
class of mechanisms, if implemented straightforwardly, does not give positive-valued RFs but, as we explain in Section \ref{sec:positivity}, can be easily transformed to variants providing positivity.

\subsection{Discretely-induced random features (DIRFs)}
Take a discrete probabilistic distribution $p(\omega)$ where $\omega_1, \dots, \omega_d$ are i.i.d. with $\mathbb{P}(\omega_l=k)=p_{k}$, $\sum_{k=0}^{\infty}p_{k}=1$ and $p_{k} > 0$ for $k \in \{ 0 \} \cup \mathbb{N}$.
Note that, by Taylor series expansion of $\exp(\cdot)$, 
\begin{equation}
K(\mathbf{x},\mathbf{y})\exp(\frac{\|\mathbf{x}\|^{2}}{2})\exp(\frac{\|\mathbf{y}\|^{2}}{2})=\exp(\mathbf{x}^{\top}{\mathbf{y}})
= \prod_{l=1}^{d}\sum_{k=0}^{\infty}p_{k}\frac{\*x_{l}^{k}\*y_{l}^{k}}{p_{k}k!}
=\mathbb{E} \left[\prod_{l=1}^{d}X_{l}\prod_{l=1}^{d}Y_{l} \right],
\end{equation}
where 
$X_{l} = \*x_{l}^{\omega_{l}}(\omega_{l}!)^{-\frac{1}{2}}p_{\omega_{l}}^{-\frac{1}{2}}, Y_{l} = \*y_{l}^{\omega_{l}}(\omega_{l}!)^{-\frac{1}{2}}p_{\omega_{l}}^{-\frac{1}{2}}$. 
Thus we can define \textit{discretely-induced random features} providing Gaussian kernel estimation as follows:
\begin{equation}
\label{dirfs}
f_{\mathrm{DI}}^{(1)}(\omega, \mathbf{x}) = f_{\mathrm{DI}}^{(2)}(\omega, \mathbf{x}) = f_{\mathrm{DI}}(\omega, \mathbf{x}) = \exp(-\frac{\|\mathbf{x}\|^{2}}{2}) \prod_{l=1}^{d}x_{i}^{\omega_{l}}(\omega_{l}!)^{-\frac{1}{2}}p_{\omega_{l}}^{-\frac{1}{2}}. 
\end{equation}
Different instantiations of the above mechanism are given by different probabilistic distributions $\{ p_k\}$. 
We will consider two prominent special cases: (a) Poisson, and (b) geometric distributions. 

\subsubsection{Poisson random features (PoisRFs)}
If $\{ p_k \}$ is a Poisson distribution, i.e. $p_k = e^{- \lambda} \lambda^k / k!$, $k \in \{ 0 \} \cup \mathbb{N}$, then the corresponding RFs are defined as: 
$f^{(1)}_\mathrm{pois} (\omega, \*x) = f^{(2)}_\mathrm{pois} (\omega, \*x) = f_\mathrm{pois} (\omega, \*x) = e^{\lambda d / 2 - \| \*x \|^2 / 2} \prod_{l = 1}^d \*x_l^{\omega_l} \lambda^{- \omega_l / 2}$. 
\begin{theorem} \label{lemma:pois} Variance of (\ref{eq:rfappr}) with $p_\mathrm{pois}, f_\mathrm{pois}$ is given by
\begin{equation}
    \mathrm{Var}_{p_\mathrm{pois} (\omega)} \left( f_\mathrm{pois} (\omega, \*x) f_\mathrm{pois} (\omega, \*y) \right) \! = \! \exp \left( \lambda d \! + \! \lambda^{-1} \sum_{l = 1}^d \*x_l^2 \*y_l^2 - \| \*x \|^2 - \| \*y \|^2 \! \right) \! - \! K(\*x, \*y)^2 . \label{eq:poisvar}
\end{equation}
\end{theorem}
The $\exp$ argument in (\ref{eq:poisvar}) is convex as a function of $\lambda > 0$. By setting its derivative to zero, we find that $\lambda^* = d^{- 1 / 2}(\sum_{l = 1}^d \*x_l^2 \*y_l^2)^{1 / 2}$ gives the minimum of (\ref{eq:poisvar}).


When, instead of a single pair $\*x, \*y$, sets $\{ \*x^{(i)} \}$, $\{ \*y^{(j)} \}$ are provided, we can use the same homogeneity assumption as in Section \ref{sec:gexp} and substitute the average of $\sum_{l = 1}^d (\*x^{(i)}_l)^2 (\*y^{(j)}_l)^2$ over $1 \leq i, j \leq L$ instead of $\sum_{l = 1}^d \*x_l^2 \*y_l^2$. This average can be computed efficiently in $O (L d)$ time as follows:
\begin{equation}
    L^{-2} \sum_{i = 1}^L \sum_{j = 1}^L \sum_{l = 1}^d (\*x^{(i)}_l)^2 (\*y^{(j)}_l)^2 = L^{-2} \sum_{l = 1}^d \left( \sum_{i = 1}^L (\*x^{(i)}_l)^2 \right) \left( \sum_{i = 1}^L (\*y^{(i)}_l)^2 \right) . \label{eq:poisstat}
\end{equation}
After computing this statistic, we can calculate $\lambda^*$ in $O (1)$ time using the analytic formula.

\subsubsection{Geometric random features (GeomRFs)}
If $\{ p_k \}$ is a geometric distribution, i.e. $p_k = p (1 - p)^k$, $k \in \{ 0 \} \cup \mathbb{N}$, for a parameter $0 < p < 1$, then the corresponding RFs are defined as: $    f^{(1)}_\mathrm{geom} (\omega, \*x) = f^{(2)}_\mathrm{geom} (\omega, \*x) = f_\mathrm{geom} (\omega, \*x) = p^{-d / 2} e^{- \| \*x \|^2 / 2} \prod_{l = 1}^d \*x_l^{\omega_l} (1 - p)^{- \omega_l / 2} (\omega_l!)^{-1 / 2}$. 
\begin{theorem} \label{lemma:geom} The variance of (\ref{eq:rfappr}) with $p_\mathrm{geom}, f_\mathrm{geom}$ is given as
\begin{equation}
    \mathrm{Var}_{p_\mathrm{geom} (\omega)} \! \left( f_\mathrm{geom} (\omega, \! \*x) f_\mathrm{geom} (\omega, \*y) \right) \! = \! p^{- d} e^{- \| \*x \|^2 - \| \*y \|^2} \prod_{l = 1}^d \! I_0 (2 (1 - p)^{- \frac12} | \*x_l \*y_l |)  - \! K(\*x, \*y)^2 \label{eq:geomvar1} 
\end{equation}
where $I_0 (\cdot)$ is the modified Bessel function of the first kind of order $0$.
\end{theorem}


Again as for the previously described mechanisms, when sets $\{ \*x^{(i)} \}$, $\{ \*y^{(j)} \}$ are given, we can use averages of $| \*x_l^{(i)} \*y_l^{(j)} |$, $1 \leq l \leq d$, instead of $| \*x_l \*y_l |$ in (\ref{eq:geomvar1}) assuming homogeneity of $\*x^{(i)}$'s and $\*y^{(j)}$'s. Each out of $d$ averages can be computed in $O (L)$ time as follows:
\begin{equation}
    L^{-2} \sum_{i = 1}^L \sum_{j = 1}^L | \*x^{(i)}_l \*y^{(j)}_l | = L^{-2} \left( \sum_{i = 1}^L |\*x^{(i)}_l| \right) \left( \sum_{i = 1}^L |\*y^{(i)}_l| \right) . \label{eq:geomstat}
\end{equation}
After precomputation of these statistics, evaluation of (\ref{eq:poisvar}) takes $O (d)$ time. A numerical optimization can be used to minimize (\ref{eq:poisvar}) with respect to $p$. As long as the number of variance evaluations is $O(L M (1 + n / d))$, the total complexity estimate is not affected.


\subsubsection{Making discretely-induced RFs positive}
\label{sec:positivity}

As can be inferred from Eq. \ref{dirfs}, DIRFs are positive when all elements of $\*x$ and $\*y$ are positive. If this is not the case, and positive-valued RFs are needed, e.g. in applications involving scalable Transformers, one way to make them positive is to take some vector $\*c \in \mathbb{R}^d$ such that $\*c_l < \*x_l, \*y_l$. An example of such a vector is given by $c_l = \min_{i} \min (\*x^{(i)}_l, \*y^{(i)}_l) - \epsilon$ where $\epsilon > 0$ is a small constant. Next, define $\widehat{\*x}^{(i)} = \*x^{(i)} - \*c$, $\widehat{\*y}^{(j)} = \*y^{(j)} - \*c$. Then, clearly, $\widehat{\*x}^{(i)} - \widehat{\*y}^{(j)} = \*x^{(i)} - \*y^{(j)}$, $K (\widehat{\*x}^{(i)}, \widehat{\*y}^{(j)}) = K (\*x^{(i)}, \*y^{(j)})$ and RFs can be used on $\widehat{\*x}^{(i)}, \widehat{\*y}^{(j)}$ which have positive entries. We refer to these variants of PoisRFs and GeomRFs as PoisRF+ and GeomRF+ respectively.

\section{Additional theoretical results \& FAVOR++}

Interestingly, as in the case of the PosRF mechanism from \cite{performer}, OPRFs also benefit from applying block-orthogonal ensembles of projections $\omega$ (see Appendix \ref{sec:ort} and \cite{performer} for the exact definition). We show below that orthogonal RFs reduce the variance of OPRFs for any $d>0$:

\begin{theorem}[Orthogonal OPRFs] \label{thm-orthoprfs}
If $\mathrm{Var}(\widehat{K}_{M}^{\mathrm{ort}}(\mathbf{x},\mathbf{y}))$ denotes the variance of the orthogonal OPRF estimator $\widehat{K}_{M}^{\mathrm{ort}}(\mathbf{x},\mathbf{y})$ of the Gaussian kernel at $\mathbf{x},\mathbf{y} \in \mathbb{R}^{d}$ using $M$ RFs and $\mathrm{Var}(\widehat{K}_{M}^{\mathrm{iid}}(\mathbf{x},\mathbf{y}))$ stands for the analogous expression but with i.i.d. samples, then for some $C(\|\mathbf{x}+\mathbf{y}\|) \geq 0$:
\begin{equation}
\mathrm{Var}(\widehat{K}_{M}^{\mathrm{ort}}(\mathbf{x},\mathbf{y})) \leq \mathrm{Var}(\widehat{K}_{M}^{\mathrm{iid}}(\mathbf{x},\mathbf{y})) - (1-\frac{1}{M})\frac{2}{d+2}C(\|\mathbf{x}+\mathbf{y}\|).
\end{equation}
\end{theorem}

\textbf{Note:} The analogous inequality can be obtained for TrigRFs only in the asymptotic sense (for $d$ large enough, see Theorem 3.8 in \cite{grfo}). One of the key properties used in the proof of Theorem \ref{thm-orthoprfs} is positivity of RFs. We conclude that positive-valued RFs are particularly well suited for the quasi Monte-Carlo methods based on the orthogonal ensembles. Analogously to FAVOR+ \cite{performer}, we refer to the self-attention approximation mechanism based on orthogonal OPRFs as \textit{FAVOR++}.

We now provide strong concentration results for the OPRF-based estimators, beyond second-moment methods, critically relying on the boundedness of OPRFs. To the best of our knowledge, these are the first such results for positive-valued RFs. Denote by $\mathcal{L}$ the Legendre Transform of the random variable $Z=f_{\mathrm{GE}}^{(1)}(\omega,\mathbf{x})f_{\mathrm{GE}}^{(2)}(\omega,\mathbf{y})$ for $f_{\mathrm{GE}}^{(\cdot)}$ as in (\ref{eq:f-func}) with $A, B, C, D \in \mathbb{R}$ defining OPRFs.

\begin{figure}[h]
    \centering
    \includegraphics[width=0.9\textwidth]{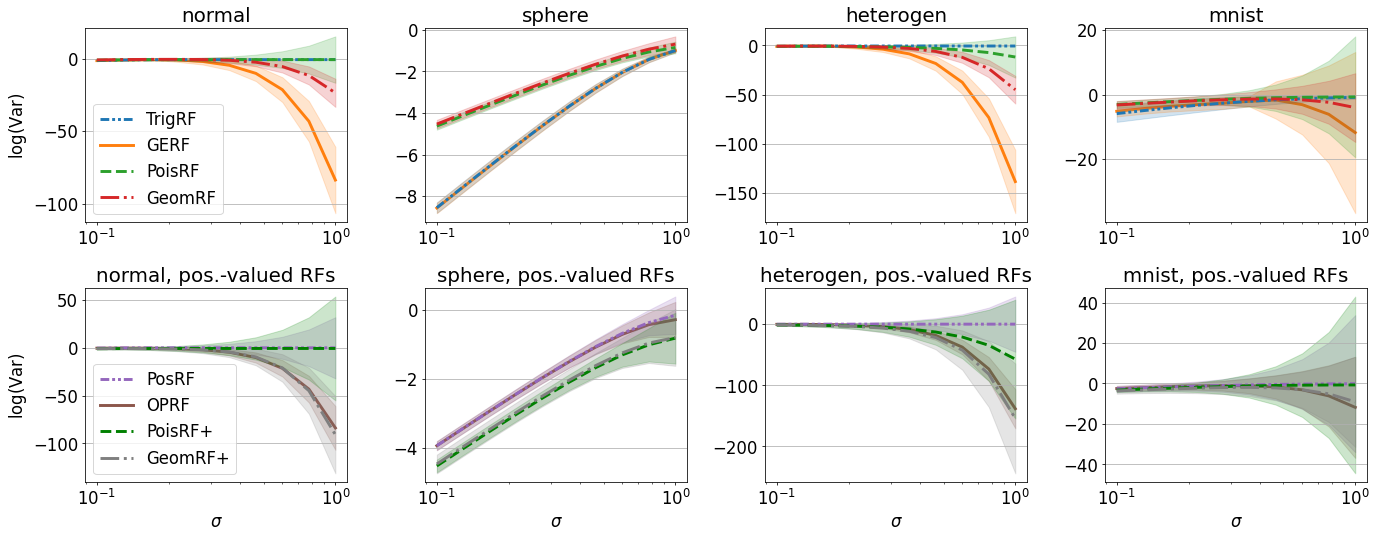}
    \caption{Log-variance of different RF mechanisms, mean and standard deviation. For each sampling method, we plot the results for non-positive and positive RFs on separate plots for $0.1 \leq \sigma \leq 1$.}
    \label{fig:log_vars}
\end{figure}

\begin{theorem}
\label{thm:tails}
The following is true for any $\epsilon>0$: $\mathbb{P}[|\widehat{K}_{M}^{\mathrm{iid}}(\mathbf{x},\mathbf{y}) - K(\mathbf{x},\mathbf{y})| \geq \epsilon] \leq 2\exp(-\frac{M \epsilon^{2}}{2} \exp(\frac{\|\mathbf{x}\|^{2}+\|\mathbf{y}\|^{2}}{2A}))$. Furthermore, for the orthogonal variant we have: $\mathbb{P}[\widehat{K}_{M}^{\mathrm{ort}}(\mathbf{x},\mathbf{y}) - K(\mathbf{x},\mathbf{y}) \geq \epsilon] \leq \exp(-M \mathcal{L}(K(\mathbf{x},\mathbf{y})+\epsilon))$
and $\mathcal{L}(K(\mathbf{x},\mathbf{y})+\epsilon)) > 0$.
\end{theorem}

Finally, below we provide the first result regarding uniform convergence for attention approximation in the efficient 
low-rank Transformers (Section \ref{sec:transf}).

\begin{theorem}[Uniform convergence for attention approximation]
\label{thm:uniform}
Assume that rows of $\*Q$ and $\*K$ from (\ref{eq:sadef}) come from the $L_2$-ball of radius $R > 0$. Denote by $\widehat{\mathcal{K}}_\mathrm{sfm}$ the approximation of $\mathcal{K}_\mathrm{sfm}$ from (\ref{eq:sadef}) via the OPRF-mechanism using $M$ independent random projections. Then $\| \mathcal{K}_\mathrm{sfm}-\widehat{\mathcal{K}}_\mathrm{sfm}\|_{\infty} \leq \epsilon$ with any constant probability when $M = \Omega(\Gamma \frac{d}{\epsilon^{2}}\log(\frac{\gamma \rho}{\epsilon}))$, where $\Gamma = \exp(-\frac{3R^{2}}{\sqrt{d}A})$, $\rho=\sqrt{2}Rd^{-\frac{1}{4}}$ and $\gamma=\sqrt{4\Gamma(\frac{R^{2}}{\sqrt{d}}+d^{2})}$ (for $A$ as in the definition of the OPRFs).
\end{theorem}

\section{Experiments}
\label{sec:exps}

We present an extensive 
empirical evaluation of CRTs. Additional details and results for each experiment can be found in the Appendix \ref{app:exp}.

\subsection{Comparing variance of CRTs}

In this initial experiment, we sample synthetic pairs of vectors $\*x, \*y$ and evaluate variance of CRTs based on the analytic formulas (\ref{eq:gexpvar},\ref{eq:poisvar},\ref{eq:geomvar1}). Our goal is to check whether there are scenarios when the newly introduced RF mechanisms have smaller variance than existing TrigRF and PosRF methods.  We set $d = 64$ which is standard in e.g. Transformer applications (Section \ref{sec:transf}). We use four different regimes for drawing $\*x, \*y$: \texttt{normal} corresponds to $\*x, \*y$ sampled from $\mathcal{N} (\*0_d, \sigma^2 \*I_d)$, \texttt{sphere} corresponds to $\*x, \*y$ sampled uniformly on a sphere $\sigma \mathcal{S}^{d-1}$, \texttt{heterogen} corresponds to $\*x$ and $\*y$ sampled from two heterogeneous distributions: $\mathcal{N} (\*0_d, \sigma^2 \*I_d)$ and $\mathcal{N} (\sigma \*1_d, \sigma^2 \*I_d)$ and \texttt{mnist} corresponds to $\*x, \*y$ being random images from MNIST dataset \cite{mnist} resized to $8 \times 8$, scaled by $\sigma > 0$ and flattened.

In many scenarios (see Figure \ref{fig:log_vars}), CRTs outperform TrigRF and PosRF baselines. Among other improvements, GERF gives more than $e^{80}$, $e^{125}$, $e^{10}$ times variance reduction compared to TrigRF in \texttt{normal}, \texttt{heterogen} and \texttt{mnist} when $\sigma = 1$. OPRF and GeomRF+ give more than $e^{75}$, $e^{125}$, $e^7$ times variance reduction compared to PosRF in \texttt{normal}, \texttt{heterogen} and \texttt{mnist} when $\sigma = 1$. 




\subsection{Comparing CRTs in the non-parametric classification}

Our next experiment is a non-parametric classification where probabilities are predicted by kernel regression \cite{nadaraya,watson} with the Gaussian kernel. Training data consists of objects $\*o^{(1)}, \dots, \*o^{(L)} \in \mathbb{R}^d$ with corresponding one-hot encoded labels $\*r^{(1)}, \dots, \*r^{(L)} \in \mathbb{R}^n$. The predicted label distribution for the new object $\*o^*$ is defined as $\*r^* = \sum_{i = 1}^L K (\sigma \*o^*, \sigma \*o^{(i)}) \*r^{(i)} / \sum_{i = 1}^L K (\sigma \*o^*, \sigma \*o^{(i)})$ where $\sigma > 0$ is a hyperparameter tuned on the validation set. Using the RF approximation for the kernel as in (\ref{eq:rfappr}), we, with $O(n L M)$ preprocessing, can approximate $\*r^*$ in $O(n M)$ time per example instead of $O(nL)$ for the exact computation.

Since the predicted class is $\mathrm{argmax}_{1 \leq l \leq n} \*r^*$, we can ignore the denominator term and, therefore, use non-positive RFs. We evaluate on classification benchmarks from UCI Repository \cite{uci} (Table \ref{tab:uci}). The best results are achieved by new RF mechanisms, with GeomRF and OPRF performing particularly well. OPRF shows the best average performance, therefore our recommendation for practitioners is to opt for this method. For the same reason, we focus on the FAVOR++ variant (OPRF  with orthogonal random projections for attention approximation) in our Transformer experiments below. 


\begin{table}[t]
\small
\centering
\caption{Non-parametric classification, test accuracy (\%). $M = 128$. The \textbf{best} result, \underline{second best}.}
\label{tab:uci}
\begin{tabular}{@{}l|cc|cccccc|r@{}}
\toprule
Dataset & TrigRF \!\! & PosRF \!\! & GERF\!\! & PoisRF\!\! & GeomRF\!\! & OPRF\!\! & PoisRF+\!\! & GeomRF+\!\! & $L$ \\
\midrule
\texttt{abalone} \cite{abalone} \!\! & $12.0$ & $16.0$ & $17.0$ & $\underline{18.0}$ & $\boldsymbol{18.3}$ & $17.1$ & $14.0$ & $15.1$ & 3758 \\
\texttt{banknote} \cite{banknote} \!\!\!\! & $66.2$ & $83.4$ & $92.4$ & $84.4$ & $\boldsymbol{94.5}$ & $\underline{92.6}$ & $80.1$ & $85.6$ & 1233 \\
\texttt{car} \cite{car} \!\! & $66.3$ & $69.2$ & $\boldsymbol{70.9}$ & $66.3$ & $66.3$ & $\underline{69.5}$ & $66.3$ & $67.2$ & 1554 \\
\texttt{yeast} \cite{yeast} \!\! & $29.7$ & $34.4$ & $\underline{42.9}$ & $36.9$ & $35.9$ & $\boldsymbol{44.4}$ & $29.7$ & $31.0$ & 1334 \\
\texttt{cmc} \cite{cmc} \!\! & $46.6$ & $45.1$ & $\boldsymbol{47.8}$ & $46.6$ & $\underline{47.3}$ & $46.3$ & $35.5$ & $43.5$ & 1324 \\
\texttt{nursery} \cite{nursery} \!\! & $31.3$ & $\underline{77.4}$ & $63.8$ & $77.1$ & $77.1$ & $\boldsymbol{78.9}$ & $77.3$ & $71.0$ & 11664 \\
\texttt{wifi} \cite{wifi} \!\! & $15.2$ & $88.8$ & $93.3$ & $\underline{95.3}$ & $\boldsymbol{95.8}$ & $93.3$ & $77.2$ & $82.9$ & 1799 \\
\texttt{chess} \cite{chess} \!\! & $16.5$ & $20.2$ & $\underline{20.4}$ & $19.1$ & $19.5$ & $20.2$ & $19.2$ & $\boldsymbol{22.5}$ & 25249 \\
\midrule
Average & $35.5$ & $54.3$ & $56.1$ & $55.5$ & $\underline{56.8}$ & $\boldsymbol{57.8}$ & $49.9$ & $52.3$ & N/A \\
\bottomrule
\end{tabular}
\end{table}

\begin{table}[b]\vspace{-6mm}
\small
\centering
   \caption{GLUE Dev results on base sized models. 
   Number of training examples is reported below each task.
   MCC score is reported for CoLA, F1 score is reported for MRPC, Spearman correlation is reported for STS-B, and accuracy scores are reported for the other tasks. The \textbf{best} result, \underline{second best}.}
   \label{tab:glue_dev}
 \begin{tabular}{@{}lcccccccc@{}}
    \toprule
System             &  MNLI & QQP  & QNLI & SST-2 & CoLA & STS-B & MRPC & RTE  \\
                   & 392k         & 363k & 108k & 67k   & 8.5k & 5.7k  & 3.5k & 2.5k   \\ 
\midrule
FAVOR+\cite{performer} & 80.26 & 89.53 & 87.13 & 90.58 & 53.17 & 85.07 & 83.82 & 67.59\\
ELU\cite{pmlr-v119-katharopoulos20a} & 80.72 & 90.05 & 89.09 & 91.51 & 48.43 & $\underline{86.68}$ & 85.05 & \textbf{68.59}\\
ReLU\cite{performer} & $\underline{81.39}$ & 90.11 & 88.85 & 91.97 & 52.08 & \textbf{87.64} & 84.56 & 67.51 \\
\midrule
FAVOR++ & 81.25 & $\underline{90.15}$ & $\underline{89.58}$ & $\underline{92.00}$ & $\underline{54.95}$ & 85.62 & $\underline{85.78}$ & $\underline{67.87}$\\
Uptrain FAVOR++ & \textbf{82.29} & \textbf{90.43} & \textbf{89.73} & \textbf{92.20} & \textbf{58.85} & 85.90 & \textbf{88.73} & 67.63\\
    \bottomrule
   \end{tabular}
\end{table}
\subsection{FAVOR++ in scalable Transformers}

\subsubsection{Natural language processing}
In this setting, we test different low-rank attention Transformers on the General Language Understanding Evaluation (GLUE) benchmark \cite{wang2018glue}, consisting of 8 different natural language understanding tasks with the sequence length ranging from 32 to 128. 
We used the same training parameters as mentioned in \cite{devlin2018bert} (see Appendix \ref{sec:appedinx_text} for details).
We compared FAVOR+ \cite{performer}, ELU \cite{pmlr-v119-katharopoulos20a} and ReLU \cite{performer} variants of the Performers \cite{performer} against a FAVOR++ variant and report the results in Table \ref{tab:glue_dev}. We find that FAVOR++ outperforms all these low-rank Transformers in most GLUE tasks. 
{In particular, FAVOR++ outperforms FAVOR+ on all GLUE tasks, demonstrating downstream effectiveness of the variance reduction of the softmax kernel estimation.}
Furthermore, warm-starting with pre-trained BERT-base model checkpoint \cite{devlin2018bert} (\emph{Uptrain FAVOR++} in Table \ref{tab:glue_dev}), further improves performance demonstrating backward-compatibility of FAVOR++ with the exact softmax kernel. 


\subsubsection{Speech modelling}
\begin{figure}[t]
\vspace{-2.5mm}
  \centering
\includegraphics[width=0.45\textwidth]{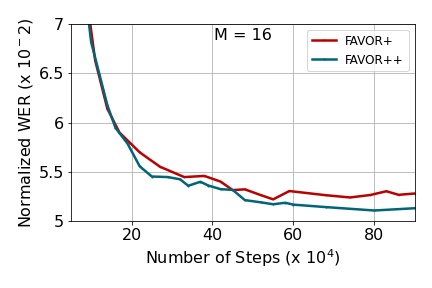} $\quad$
  \includegraphics[width=0.45\textwidth]{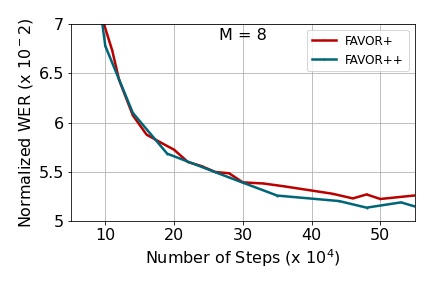}
  \vspace{-3.5mm}
  \caption{\small{Comparison of the Conformer-Transducer encoder with FAVOR++ and FAVOR+ attention on the LibriSpeech \cite{librispeech} corpus for $M=16$ and $M=8$ RFs. We used common word error rate (WER) metric.}}
  \label{fig:speech}
\vspace{-1.5mm}
\end{figure}

We compare FAVOR++ with FAVOR+ on speech models with the LibriSpeech ASR corpus (\cite{librispeech}). We apply both to approximate attention blocks in the $17$-layer Conformer-Transducer encoder (\cite{conformer}) of only 4 attention heads and use the word error rate (WER) metric -- a standard way to evaluate speech models.
In both cases FAVOR++ outperforms FAVOR+, as shown in Figure \ref{fig:speech}. The WER improvement for FAVOR++ is substantial: 2.49\% for $M = 8$ and 3.05\% for $M = 16$ with a negligible $O (L d) \ll O(L M d)$ overhead for computing (\ref{eq:gestat}) compared to FAVOR+.

\vspace{-2mm}
\subsubsection{Vision Transformers}
\begin{figure}[t]
\vspace{-2.5mm}
  \centering
  \includegraphics[width=0.45\textwidth]{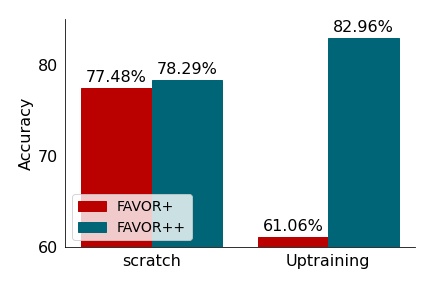} $\quad$
  \includegraphics[width=0.45\textwidth]{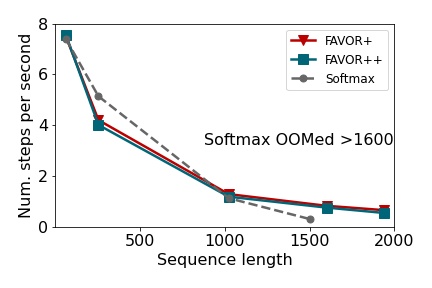}
  \vspace{-4.5mm}
  \caption{\small{Image-Transformers experiments. \textbf{Left:} Accuracy of training FAVOR+ and FAVOR++ on ImageNet from scratch and fine-tuning from softmax MAE pre-trained weights (Uptraining). \textbf{Right:} Comparing sequence length vs number of steps per second for FAVOR+, FAVOR++ and regular Transformer variant ($\mathrm{Softmax}$).}}
  \label{fig:img}
\vspace{-1.5mm}
\end{figure}

To further showcase the need for more accurate softmax kernel approximation, we compare the performance of FAVOR+ and FAVOR++ on ImageNet (\cite{imagenet}). We inject both mechanisms to the attention modules of Vision Transformers (ViT \cite{dosovitskiy2020image}). 
In Figure \ref{fig:img}, we show the results of training from scratch and uptraining from the 
MAE checkpoint \cite{he2021masked}. 
We see that, as opposed to FAVOR+, FAVOR++ is more stable and is able to improve performance especially for uptraining, demonstrating backward-compatibility with the exact softmax kernel. 

Finally, we compare the computational complexity of FAVOR+ and FAVOR++. In Figure \ref{fig:img}, the right plot shows the number of steps per second as a function of sequence length $L$ on the same hardware. We see that attention modules using FAVOR+ and FAVOR++ have very similar  computation time (both provide linear attention). Moreover, for sequence lengths above $1000$, training a regular ViT model became increasingly difficult due to out-of-memory errors. 

\section{Limitations of this work \& broader impact}\label{sec:lim}


Several of the mechanisms proposed in this paper can be further extended, potentially leading to even more accurate algorithms. For instance, it remains an open question how to choose theoretically optimal parameters for GERF and GeomRF mechanisms. Furthermore, DIRFs can benefit from optimizing the discrete distributions defining them (to minimize variance of the estimation) rather than choosing them a priori. Our methods should be used responsibly, given rising concerns regarding the carbon footprint of training massive Transformer models and other societal issues \cite{co2,gptleak,stochpar,weidinger2021ethical}. 

\section{Conclusion}
\label{sec:conclusion}

We presented a new class of RF mechanisms called chefs' random tables (CRTs) including methods providing positivity and boundedness of random features -- two key properties for new applications of RFs in Transformer training. We provided comprehensive theoretical results and extensive empirical evaluation, resulting in particular in new state-of-the-art low-rank attention Transformers for text.

\section{Acknowledgements}

V. L. acknowledges support from the Cambridge Trust and DeepMind. V. L. was part-time employed by Google while a PhD student. A.W. acknowledges support from a Turing AI Fellowship under EPSRC grant EP/V025279/1, The Alan Turing Institute, and the Leverhulme Trust via CFI.


\bibliographystyle{plain}
\bibliography{references}



\newpage
{
\begin{center}
    \Large
    \textbf{{Chefs' Random Tables: Non-Trigonometric Random Features -- Appendix}}
\end{center}
}

\section{Appendix}


\subsection{Orthogonal random projections}
\label{sec:ort}
The orthogonal random projections mechanism (\cite{performer}) is the Monte Carlo method, where samples $\omega_{1},...,\omega_{M}$, marginally distributed as $\mathcal{N}(\*0_d,\mathbf{I}_{d})$ (thus maintaining unbiasedness of the overall mechanism), are conditioned to form an orthogonal ensemble when $M \leq d$, otherwise samples are partitioned into $d \times d$ independent orthogonal blocks. Orthogonal random projections can be easily constructed form th iid projections via the Gram-Schmidt orthogonalization algorithm (see: \cite{unrea}).

\subsection{Proof of Theorem \ref{lemma:gerfs}} 
\label{app:gerfs}

\begin{proof}
We rewrite (\ref{eq:rf}) for $f^{(\cdot)}_\mathrm{GE}$ and deduce that
\begin{align}
    &\mathbb{E} \left( f^{(1)}_\mathrm{GE} (\omega, \*x) f^{(2)}_\mathrm{GE} (\omega, \*y) \right) = (2 \pi)^{- d / 2} D^2 \int_{\mathbb{R}^d} \exp (- \| \omega \|^2 / 2 + 2 A \| \omega \|^2 + B \omega^\top (\*x + s \*y) \nonumber \\
    &+ C (\|\*x\|^2 + \|\*y\|^2)) d \omega \nonumber \\
    &= (2 \pi)^{- d / 2} D^2 \exp \left( \frac{B^2}{2 (1 - 4 A)} \|\*x + s \*y\|^2 + C (\| \*x \|^2 + \| \*y \|^2) \right) \nonumber \\
    &\times \int_{\mathbb{R}^d} \exp \biggl(- \frac12 (1 - 4 A) \left[ \omega - \frac{B}{1 - 4 A} (\*x + s \*y) \right]^2 \biggr) d \omega \nonumber \\
    &= D^2 \left(\sqrt{1 - 4 A}\right)^{-d} \exp \left( \frac{B^2}{2 (1 - 4 A)} \|\*x + s \*y\|^2 + C (\| \*x \|^2 + \| \*y \|^2) \right) \nonumber \\
    &= D^2 \left(\sqrt{1 - 4 A}\right)^{-d} \exp \left( \frac{s B^2}{(1 - 4 A)} \*x^\top \*y + \left( \frac{B^2}{2 (1 - 4 A)} + C \right) (\| \*x \|^2 + \| \*y \|^2) \right)
    \label{eq:cint1}
\end{align}
where by $[\cdot]^2$ we denote an elementwise square of the input vector and we use an identity:
\begin{equation}
    \int_{\mathbb{R}^d} \exp \biggl(- \frac{\alpha}{2} \left[\omega  - \beta \right]^2 \biggr) d \omega = (2 \pi)^{d / 2} \left(\sqrt{\alpha}\right)^{-d} . \label{eq:cgauss}
\end{equation}
where $\alpha \in \mathbb{C}$, $\re{\alpha} > 0$ ($\alpha = 1 - 4 A$ in (\ref{eq:cint1})) and $\beta \in \mathbb{C}^d$ ($\beta = (B / (1 - 4 A)) (\*x + s \*y)$ in (\ref{eq:cint1})). When both $\alpha$ and $\beta$ are real, (\ref{eq:cgauss}) is correct since it is integral of the scaled multivariate Gaussian density. Since both the left and the right hand side in (\ref{eq:cgauss}) are analytic functions of $\alpha$ and $\beta$ when $\re{\alpha} > 0$, by the identity theorem \cite{identity} we conclude that (\ref{eq:cgauss}) holds when $\alpha$ and $\beta$ are complex and $\re{\alpha} > 0$.

The right hand side of (\ref{eq:cint1}) is $K (\*x, \*y)$ if the following conditions are satisfied in addition to $\re{1 - 4 A} > 0$:
\begin{equation}
    D^2 = (\sqrt{1 - 4 A})^d, \quad s B^2 = (1 - 4 A), \quad \frac{B^2}{2 (1 - 4 A)} + C = -\frac12 . \label{eq:abcd2}
\end{equation}
(\ref{eq:abcd2}) is satisfied when (\ref{eq:abcd}) takes place. The final thing to mention is that $\re{\cdot}$ is a linear operation, and therefore, if (\ref{eq:abcd2}) is satisfied,
\begin{equation*}
    \mathbb{E} \re{f^{(1)}_\mathrm{GE} (\omega, \*x) f^{(2)}_\mathrm{GE} (\omega, \*y)} = \re{\mathbb{E} \left(f^{(1)}_\mathrm{GE} (\omega, \*x) f^{(2)}_\mathrm{GE} (\omega, \*y) \right)} = \re{K (\*x, \*y)} = K (\*x, \*y).
\end{equation*}
\end{proof}
It's possible to use other complex roots in (\ref{eq:abcd}) rather than just principal roots. However, in the proof of Theorem \ref{lemma:gevar}, we will only use (\ref{eq:abcd2}) and, therefore, the variance is the same when other complex roots are used. We opt for principal roots for simplicity.

\subsection{Proof of Theorem \ref{lemma:gevar}}
\label{app:gevar}

\begin{proof}
Denote $f^{(1)}_\mathrm{GE}$ and $f^{(2)}_\mathrm{GE}$ with parameters $A, B, C, D$ as $f^{(1)}_{A,B,C,D}, f^{(2)}_{A,B,C,D}$. Then
\begin{align}
    \mathrm{Var}_{p_\mathrm{GE}}& \re{f^{(1)}_\mathrm{GE} (\omega, \*x) f^{(2)}_\mathrm{GE} (\omega, \*y)} = \mathbb{E} \, \re{f^{(1)}_\mathrm{GE} (\omega, \*x) f^{(2)}_\mathrm{GE} (\omega, \*y)}^2 \nonumber \\
    &- \left( \mathbb{E} \, \re{f^{(1)}_\mathrm{GE} (\omega, \*x) f^{(2)}_\mathrm{GE} (\omega, \*y)} \right)^2 \nonumber \\
    &= \frac14 \mathbb{E} \left( f^{(1)}_\mathrm{GE} (\omega, \*x) f^{(2)}_\mathrm{GE} (\omega, \*y) + \overline{f^{(1)}_\mathrm{GE} (\omega, \*x)} \overline{f^{(2)}_\mathrm{GE} (\omega, \*y)} \right)^2 - K (\*x, \*y)^2 \nonumber \\
    &= \frac14 \mathbb{E} \left( f^{(1)}_\mathrm{GE} (\omega, \*x)^2 f^{(2)}_\mathrm{GE} (\omega, \*y)^2 \right) + \frac12 \mathbb{E} \left( | f^{(1)}_\mathrm{GE} (\omega, \*x) |^2 | f^{(2)}_\mathrm{GE} (\omega, \*y) |^2 \right) \nonumber \\
    &+ \frac14 \overline{\mathbb{E} \left( f^{(1)}_\mathrm{GE} (\omega, \*x)^2 f^{(2)}_\mathrm{GE} (\omega, \*y)^2 \right)} - K (\*x, \*y)^2 \nonumber \\
    &= \frac12 \re{\mathbb{E} \left( f^{(1)}_\mathrm{GE} (\omega, \*x)^2 f^{(2)}_\mathrm{GE} (\omega, \*y)^2 \right)} + \frac12 \mathbb{E} \left( | f^{(1)}_\mathrm{GE} (\omega, \*x) |^2 | f^{(2)}_\mathrm{GE} (\omega, \*y) |^2 \right) - K (\*x, \*y)^2 \nonumber \\
    &= \frac12 \re{\mathbb{E} \left( f^{(1)}_{2 A, 2 B, 2 C, D^2} (\omega, \*x) f^{(2)}_{2 A, 2 B, 2 C, D^2} (\omega, \*y) \right)} \nonumber \\
    &+ \frac12 \mathbb{E} \left( f^{(1)}_{2 \re{A}, 2 \re{B}, 2 \re{C}, | D |^2} (\omega, \*x) f^{(2)}_{2 \re{A}, 2 \re{B}, 2 \re{C}, | D |^2} (\omega, \*y) \right) \nonumber \\
    &- K (\*x, \*y)^2. \label{eq:varder1}
\end{align}
In the new notation, (\ref{eq:cint1}) is $\mathbb{E} \left( f^{(1)}_{A,B,C,D} (\omega, \*x) f^{(2)}_{A,B,C,D} (\omega, \*y) \right)$. By substituting $A, B, C, D \to 2 A, 2 B, 2 C, D^2$ and $A, B, C, D \to 2 \re{A}, 2 \re{B}, 2 \re{C}, | D |^2$ into (\ref{eq:cint1}) (it's possible since $\re{1 - 8 A} > 0$ and, hence, $\re{1 - 8 \re{A}} > 0$), we can compute expectations in (\ref{eq:varder1}):
\begin{align}
    \mathbb{E} &\left( f^{(1)}_{2 A, 2 B, 2 C, D^2} (\omega, \*x) f^{(2)}_{2 A, 2 B, 2 C, D^2} (\omega, \*y) \right) = D^4 \left(\sqrt{1 - 8 A}\right)^{-d} \nonumber \\
    &\cdot \exp \biggl( \frac{4 B^2}{2 (1 - 8 A)} \|\*x + s \*y\|^2 + 2 C (\| \*x \|^2 + \| \*y \|^2) \biggr) \nonumber \\
    &= \left(\sqrt{\frac{(1 - 4 A)^2}{1 - 8 A}}\right)^d \exp \biggl( \frac{4 s (1 - 4 A)}{2 (1 - 8 A)} \|\*x + s \*y\|^2 - (s + 1) (\| \*x \|^2 + \| \*y \|^2) \biggr) \nonumber \\
    &= \alpha_1 \exp \left( \alpha_2 \|\*x + s \*y\|^2 - (s + 1) \left( \| \*x \|^2 + \| \*y \|^2 \right) \right), \label{eq:varder2} \\
    \mathbb{E} &\left( f^{(1)}_{2 \re{A}, 2 \re{B}, 2 \re{C}, | D |^2} (\omega, \*x) f^{(2)}_{2 \re{A}, 2 \re{B}, 2 \re{C}, | D |^2} (\omega, \*y) \right) = | D |^4 \left(\sqrt{1 - 8 \re{A}} \right)^{-d} \nonumber \\
    &\cdot \exp \biggl( \frac{(B + \overline{B})^2}{2 (1 - 8 \re{A})} \|\*x + s \*y\|^2 + 2 \re{C} (\| \*x \|^2 + \| \*y \|^2) \biggr) \nonumber \\
    &= \left(\frac{(1 - 4 A) \overline{(1 - 4 A)}}{1 - 8 \re{A}} \right)^{d / 2} \exp \biggl( \frac{B^2 + \overline{B^2} + 2 | B |^2}{2 (1 - 8 \re{A})} \|\*x + s \*y\|^2 - (s + 1) (\| \*x \|^2 + \| \*y \|^2) \biggr) \nonumber \\
    &= \left(\frac{1 - 8 \re{A} + 16 | A |^2}{1 - 8 \re{A}} \right)^{d / 2} \nonumber \\
    &\cdot \exp \biggl( \frac{s(2 - 8 \re{A}) + 2 | 1 - 4 A |}{2 (1 - 8 \re{A})} \|\*x + s \*y\|^2 - (s + 1) (\| \*x \|^2 + \| \*y \|^2) \biggr) \nonumber \\
    &= \alpha_3 \exp \left( \alpha_4 \|\*x + s \*y\|^2 - (s + 1) \left( \| \*x \|^2 + \| \*y \|^2 \right) \right) \label{eq:varder3}
\end{align}
where we use (\ref{eq:abcd2}) to express $B^2, C, D^2$ through $A$ and $C$. (\ref{eq:varder1},\ref{eq:varder2},\ref{eq:varder3}) together result in (\ref{eq:gexpvar}).
\end{proof}

\subsection{Proof of Theorem \ref{lemma:minvar}}
\label{app:minvar}

\begin{proof}
When $A$ is real and $s = +1$, variance (\ref{eq:gexpvar}) has a form:
\begin{align*}
    \mathrm{Var}_{p_\mathrm{GE}}& \left( f^{(1)}_\mathrm{GE} (\omega, \*x) f^{(2)}_\mathrm{GE} (\omega, \*y) \right) = \left(\frac{1 - 4 A}{\sqrt{1 - 8 A}}\right)^d \\
    &\cdot \exp \biggl( \frac{2 (1 - 4 A)}{1 - 8 A} \|\*x + \*y\|^2 - 2 (\| \*x \|^2 + \| \*y \|^2) \biggr) - K (\*x, \*y)^2 \\
    &= 2^{-d} \left(\frac{\rho + 1}{\sqrt{\rho}} \right)^d \exp \left(\left(1 + \rho \right) \| \*x + \*y \|^2 - 2 (\|\*x\|^2 + \|\*y\|^2)\right) - K (\*x, \*y)^2
\end{align*}
where we change the variable $\rho = \frac{1}{1 - 8 A} \in (0, +\infty)$. We see that the minimum of variance with respect to $\rho \in (0, +\infty)$ coincides with the minimum of the logarithm of the first term:
\begin{equation*}
    g (\rho) = -d \log 2 + d \log (\rho + 1) - \frac{d}{2} \log \rho + (1 + \rho) \| \*x + \*y \|^2 - 2 (\| \*x \|^2 + \| \*y \|^2).
\end{equation*}
All stationary points $\rho^*$ can be found by setting its derivative to zero:
\begin{equation*}
    g' (\rho^*) = \frac{d}{\rho^* + 1} - \frac{d}{2 \rho^*} + \| \*x + \*y \|^2 = 0.
\end{equation*}
Multiply by $2 \rho^* (\rho^* + 1) > 0$ and obtain an equivalent quadratic equation:
\begin{gather}
    d (\rho^* - 1) + 2 \rho^* (\rho^* + 1) \| \*x + \*y \|^2 = 0; \nonumber \\
    2 \| \*x + \*y \|^2 (\rho^*)^2 + (2 \| \*x + \*y \|^2 + d) \rho^* - d = 0; \nonumber \\
    \rho^*_{1,2} = \frac{1}{4 \| \*x + \*y \|^2} \left( \pm \sqrt{(2 \| \*x + \*y \|^2 + d)^2 + 8 d \| \*x + \*y \|^2} - 2 \| \*x + \*y \|^2 - d \right) . \label{eq:roots}
\end{gather}
The root $\rho^*_2$ of the quadratic equation with ``$-$'' sign in place of ``$\pm$'' (\ref{eq:roots}) is a negative number. Since $\| \*x + \*y \|^2 > 0$, we conclude that the only stationary point is the positive root $\rho^* = \rho^*_1$ with ``$+$'' sign in place of ``$\pm$''.

$g' (\rho)$ is a continuous function with $g' (\rho) \to - \infty$ as $\rho \to +0$ and $g' (\rho) \to \| \*x + \*y \|^2 > 0$ as $\rho \to +\infty$. There is only one $\rho^*$ such that $g' (\rho^*) = 0$, and therefore for all $\rho < \rho^*$, $g'(\rho) < 0$ and for all $\rho > \rho^*$, $g'(\rho) < 0$. Hence, $\rho^*$ is a global minimum of $g(\rho)$.
\end{proof}

\subsection{Proof of Theorem \ref{lemma:pois}}
\label{app:pois}

\begin{proof}
Variance of the estimator has a form:
\begin{align*}
    \mathrm{Var}_{p_\mathrm{pois}} \left( f_\mathrm{pois} (\omega, \*x) f_\mathrm{pois} (\omega, \*y) \right) &= \mathbb{E} \left( f_\mathrm{pois} (\omega, \*x)^2 f_\mathrm{pois} (\omega, \*y)^2 \right) - \left( \mathbb{E} \left( f_\mathrm{pois} (\omega, \*x) f_\mathrm{pois} (\omega, \*y) \right) \right)^2 \\
    &= e^{2 \lambda d - \| \*x \|^2 - \| \*y \|^2} \prod_{l = 1}^d \mathbb{E} \left( (\*x_l \*y_l)^{2 \omega_l} \lambda^{-2 \omega_l} \right) - K (\*x, \*y)^2 \\
    &= e^{2 \lambda d - \| \*x \|^2 - \| \*y \|^2} \prod_{l = 1}^d e^{- \lambda} \sum_{k = 0}^\infty \frac{\lambda^k}{k!} (\*x_i \*y_i)^{2k} \lambda^{- 2 k} - K (\*x, \*y)^2 \\
    &= e^{\lambda d - \| \*x \|^2 - \| \*y \|^2} \prod_{l = 1}^d \sum_{k = 0}^\infty \frac{(\*x_l^2 \*y_l^2 \lambda^{-1})^k}{k!} - K (\*x, \*y)^2 \\
    &= e^{\lambda d - \| \*x \|^2 - \| \*y \|^2} \prod_{l = 1}^d \exp \left( \frac{\*x_l^2 \*y_l^2}{\lambda} \right) - K (\*x, \*y)^2 \\
    &= \exp \left( \lambda d + \lambda^{-1} \sum_{l = 1}^d \*x_l^2 \*y_l^2 - \| \*x \|^2 - \| \*y \|^2 \right) - K (\*x, \*y)^2.
\end{align*}
\end{proof}

\subsection{Proof of Theorem \ref{lemma:geom}}
\label{app:geom}

\begin{proof}
Variance of the estimator has a form:
\begin{align*}
    \mathrm{Var}_{p_\mathrm{geom}} &\left( f_\mathrm{geom} (\omega, \*x) f_\mathrm{geom} (\omega, \*y) \right) = \mathbb{E} \left( f_\mathrm{geom} (\omega, \*x)^2 f_\mathrm{geom} (\omega, \*y)^2 \right) \\
    &- \left( \mathbb{E} \left( f_\mathrm{geom} (\omega, \*x) f_\mathrm{geom} (\omega, \*y) \right) \right)^2 \\
    &= p^{-2 d} e^{- \| \*x \|^2 - \| \*y \|^2} \prod_{l = 1}^d \mathbb{E} (\omega_l !)^{-2} \left( ((1 - p)^{-1} \*x_l \*y_l)^{2 \omega_l} \right) - K (\*x, \*y)^2 \\
    &= p^{-2 d} e^{- \| \*x \|^2 - \| \*y \|^2} \prod_{l = 1}^d \sum_{k = 0}^\infty p (1 - p)^k (k!)^{-2} ((1 - p)^{-1} \*x_i \*y_i)^{2k} - K (\*x, \*y)^2 \\
    &= p^{-d} e^{- \| \*x \|^2 - \| \*y \|^2} \prod_{l = 1}^d \sum_{k = 0}^\infty (k!)^{-2} ( (1 - p)^{-1 / 2} \*x_l \*y_l )^{2k}  - K (\*x, \*y)^2 \\
    &= p^{-d} e^{- \| \*x \|^2 - \| \*y \|^2} \prod_{l = 1}^d I_0 ( 2 (1 - p)^{-1 / 2} \*x_l \*y_l )  - K (\*x, \*y)^2 \\
    &= p^{-d} e^{- \| \*x \|^2 - \| \*y \|^2} \prod_{l = 1}^d I_0 ( 2 (1 - p)^{-1 / 2} | \*x_l \*y_l | )  - K (\*x, \*y)^2
\end{align*}
where we use Taylor series $I_0 (x) = \sum_{k = 0}^\infty (k!)^{-2} (x / 2)^{2 k}$ and the even parity of $I_0 (x)$.
\end{proof}
We take absolute values $| \*x_l \*y_l |$ instead of just $\*x_l \*y_l$ because the average of $\*x_l^{(i)}$ and $\*y_l^{(j)}$ would converge to zero due to different signs and wouldn't produce any meaningful statistic.

\subsection{Proof of Theorem \ref{thm-orthoprfs}}
\label{app:orthoprfs}

We prove here a much more general result from which Theorem \ref{thm-orthoprfs} follows. 

\begin{theorem}
\label{thm-ortho}
Consider a random variable $X$ of the form: $X = g(\omega^{\top}\mathbf{z}, \|\omega\|)$ for some fixed $\mathbf{z} \in \mathbb{R}^{d}$ and: $g:\mathbb{R} \times \mathbb{R}_{\geq 0} \rightarrow \mathbb{R}$, where $\omega$ is sampled from the isotropic distribution $\Omega(d)$ with the corresponding distribution of $\|\omega\|$ denoted as $\tilde{\Omega}(d)$. Assume furthermore that for every $y \in \mathbb{R}_{\geq 0}$, function $g_{y}:\mathbb{R} \to \mathbb{R}$, defined as $g_{y}(x)=g(x,y)$, satisfies: $g_{y}(x) = \sum_{k=0}^{\infty} a_{k}(y) x^{k}$ for some $a_{0}(y),a_{1}(y),... \geq 0$. Take two unbiased estimators of $K=\mathbb{E}[X]$, defined for $M \leq d$ as:
\begin{equation}
\widehat{K}^{\mathrm{iid}}_{M} = \frac{1}{M}\sum_{m=1}^{M} g((\omega_{m}^{\mathrm{iid}})^{\top}\mathbf{z},\|\omega_{m}\|), \textrm{      } \widehat{K}^{\mathrm{ort}}_{M} = \frac{1}{M}\sum_{m=1}^{M} g((\omega_{m}^{\mathrm{ort}})^{\top}\mathbf{z},\|\omega_{m}\|) 
\end{equation}
for $\omega_{1}^{\mathrm{iid}},...,\omega_{M}^{\mathrm{iid}} \overset{\mathrm{iid}}{\sim} \mathcal{N}(\*0_d,\mathbf{I}_{d})$ and
the orthogonal ensemble $\omega_{1}^{\mathrm{ort}},...,\omega_{M}^{\mathrm{ort}} \sim \mathcal{N}(\*0_d,\mathbf{I}_{d})$, then:
\begin{equation}
\mathrm{Var}(\widehat{K}^{\mathrm{ort}}_{M}) \leq \mathrm{Var}(\widehat{K}^{\mathrm{iid}}_{M}) - (1-\frac{1}{M})\frac{2}{d+2}F^{2}(\mathbf{z}),
\end{equation}
where $F(\mathbf{z}) \overset{\mathrm{def}}{=} \mathbb{E}_{\mathbf{u} \sim \mathrm{Unif}(0,\mathcal{S}^{d-1})}\mathbb{E}_{x \sim \tilde{\Omega}(d)}\left[\tilde{g}(\mathbf{u}^{\top}\mathbf{z}, x)\right]$,
$\mathrm{Unif}(0,\mathcal{S}^{d-1})$ is the uniform probabilistic distribution on the $(d-1)$-dimensional unit sphere in $\mathbb{R}^{d}$ and $\tilde{g}(a,b) \overset{\mathrm{def}}{=} \frac{g(a,b)+g(-a,b)}{2}-g(0,b)$.
\end{theorem}
If we define $g$ as: $g(a,b)=D^{2}\exp(2Ab^{2}+Ba+2C(\|\mathbf{x}\|^{2}+\|\mathbf{y}\|^{2}))$ for $A,B,C \in \mathbb{R}$ (see: Sec. \ref{sec:gexp}), take $\Omega=\mathcal{N}(\*0_d,\mathbf{I}_{d})$ and $\mathbf{z}=\mathbf{x}+\mathbf{y}$ then $\widehat{K}^{\mathrm{iid}}_{M}$ and $\widehat{K}^{\mathrm{ort}}_{M}$  from Theorem \ref{thm-ortho} become the estimators of the Gaussian kernel applying $M$ generalized exponential random features that are either i.i.d or constructed from the orthogonal ensembles. Therefore, as a corollary we obtain Theorem \ref{thm-orthoprfs}.

\begin{proof}
We start by factorizing the variance of $K^{\mathrm{iid}}_{M}$ and $K^{\mathrm{ort}}_{M}$ by conditioning on the lengths of the used random samples. We have:
\begin{equation}
\mathrm{Var}(K^{\mathrm{iid}}_{M}) = 
\int_{\mathbb{R} \times ... \times \mathbb{R}}\mathrm{Var}\left(K^{\mathrm{iid}_{m}} \textrm{   }  | \textrm{   }  \{\|\omega_{1}^{\mathrm{iid}}\|=x_{1},...,\|\omega_{m}^{\mathrm{iid}}\|=x_{M}\}\right)\prod_{m=1}^{M} \mathcal{P}(x_{m})\cdot dx_{1} \cdot ... \cdot dx_{M},
\end{equation}
and similarly:
\begin{equation}
\mathrm{Var}(K^{\mathrm{ort}}_{M}) = 
\int_{\mathbb{R} \times ... \times \mathbb{R}}\mathrm{Var}\left(K^\mathrm{ort}_{M} \textrm{   }  | \textrm{   }  \{\|\omega_{1}^{\mathrm{ort}}\|=x_{1},...,\|\omega_{M}^{\mathrm{ort}}\| =x_{M}\}\right)\prod_{m=1}^{M} \mathcal{P}(x_{m})\cdot dx_{1} \cdot ... \cdot dx_{M},
\end{equation}
where $\mathcal{P}$ is the pdf function for the distribution $\tilde{\Omega}(d)$ of the lengths of samples taken from $\Omega(d)$.
We use the fact that in both scenarios: iid samples and an orthogonal ensemble, the lengths of vectors $\omega_{i}$ are sampled from the same distribution $\tilde{\Omega}$, independently from their directions and from each other.
Therefore we have:
\begin{equation}
\mathrm{Var}(K^{\mathrm{iid}}_{M}) - 
\mathrm{Var}(K^{\mathrm{ort}}_{M}) = 
\int_{\mathbb{R} \times ... \times \mathbb{R}} T(x_{1},...,x_{M}) \prod_{m=1}^{M} \mathcal{P}(x_{m}) \cdot dx_{1} \cdot ... \cdot dx_{M},
\end{equation}
where 
\begin{align}
\begin{split}
T(x_{1},...,x_{M}) =  \mathrm{Var}\left(K^\mathrm{iid}_{M} \textrm{   }  | \textrm{   }  \{\|\omega_{1}^{\mathrm{iid}}\|=x_{1},...,\|\omega_{M}^{\mathrm{iid}}\|=x_{M}\}\right) - \\ \mathrm{Var}\left(K^\mathrm{ort}_{M} \textrm{   }  | \textrm{   }  \{\|\omega_{1}^{\mathrm{ort}}\|=x_{1},...,\|\omega_{M}^{\mathrm{ort}}\|=x_{M}\}\right)
\end{split}
\end{align}
Since the lengths of the samples are chosen independently from their directions, we conclude that:
\begin{equation}
\mathrm{Var}\left(K^{\mathrm{iid}}_{m} \textrm{   }  | \textrm{   }  \{\|\omega_{1}^{\mathrm{iid}}\|=x_{1},...,\|\omega_{m}^{\mathrm{iid}}\|=x_{M}\}\right) = \mathrm{Var}\left(\frac{1}{M}\sum_{m=1}^{M}X^{\mathrm{iid}}_{m} \right)    
\end{equation}
and 
\begin{equation}
\mathrm{Var}\left(K^{\mathrm{ort}}_{M} \textrm{   }  | \textrm{   }  \{\|\omega_{1}^{\mathrm{ort}}\|=x_{1},...,\|\omega_{M}^{\mathrm{ort}}\|=x_{m}\}\right) = \mathrm{Var}\left(\frac{1}{M}\sum_{m=1}^{M}X^{\mathrm{ort}}_{m}\right),    
\end{equation}
where $X^{\mathrm{iid}}_{m} = g_{x_{m}}((\mathbf{u}_{m}^{\mathrm{iid}})^{\top}\mathbf{z})$ and $X^{\mathrm{ort}}_{m} = g_{x_{m}}((\mathbf{u}_{m}^{\mathrm{ort}})^{\top}\mathbf{z})$,
$\{\mathbf{u}^{\mathrm{iid}}_{1},...,\mathbf{u}^{\mathrm{iid}}_{M}\}$ are iid samples from the unit-sphere in $\mathbb{R}^{d}$ and $\{\mathbf{u}^{\mathrm{ort}}_{1},...,\mathbf{u}^{\mathrm{ort}}_{M}\}$ is an orthogonal ensemble of samples taken from the unit sphere in $\mathbb{R}^{d}$.

Thus we have: 
\begin{equation}
T(x_{1},...,x_{M}) = \mathrm{Var}(\frac{1}{M}\sum_{m=1}^{M}X^{\mathrm{iid}}_{m}) -
\mathrm{Var}(\frac{1}{M}\sum_{m=1}^{M}X^{\mathrm{ort}}_{m})
\end{equation}
Now, by the similar analysis as in the proof of Theorem 5 in \cite{performer}, we obtain for $r \sim \mathcal{N}(0,1)$:

\begin{align}
\begin{split}
T(x_{1},...,x_{M}) \geq \frac{2}{(d+2)}\cdot \frac{2}{M^{2}}\sum_{1 \leq i < j \leq M}\sum_{t,u=1}^{\infty}a_{2t}(x_{i})a_{2u}(x_{j})\|\mathbf{z}\|^{2t+2u}
 \mathbb{E}[\|\omega\|^{2t}] \mathbb{E}[\|\omega\|^{2u}] \cdot \\
\frac{\mathbb{E}[r^{2t}]\mathbb{E}[r^{2u}]}
{\mathbb{E}[\sqrt{g_{1}^{2}+...+g_{d}^{2}}^{2t}]
\mathbb{E}[\sqrt{g_{1}^{2}+...+g_{d}^{2}}^{2u}]}
= \frac{2}{d+2}\cdot \frac{2}{M^{2}} \sum_{1 \leq i < j \leq M}  \\   
\left(\sum_{t=1}^{\infty}a_{2t}(x_{i})\|\mathbf{z}\|^{2t}
  \cdot 
\frac{\mathbb{E}[\|\omega\|^{2t}] \cdot \mathbb{E}[r^{2t}]}
{\mathbb{E}[\sqrt{g_{1}^{2}+...+g_{d}^{2}}^{2t}]}\right) \cdot 
\left(\sum_{t=1}^{\infty}a_{2t}(x_{j})\|\mathbf{z}\|^{2t}
  \cdot 
\frac{\mathbb{E}[\|\omega\|^{2t}] \cdot \mathbb{E}[r^{2t}]}
{\mathbb{E}[\sqrt{g_{1}^{2}+...+g_{d}^{2}}^{2t}]}\right)\\
= \frac{2}{d+2} \cdot \frac{2}{M^{2}}\sum_{1 \leq i<j \leq M}
F_{x_{i}}(\mathbf{z})F_{x_{j}}(\mathbf{z}),
\end{split}    
\end{align}
where $F_{x}(\mathbf{z}) \overset{\mathrm{def}}{=} \mathbb{E}[\tilde{g}(\mathbf{u}^{\top}\mathbf{z},x)]$, $\tilde{g}(a,b)\overset{\mathrm{def}}{=}\frac{g(a,b)+g(-a,  b) }{2} - g(0,b)$ and $\mathbf{u} \sim \mathrm{Unif}(\mathcal{S}^{d-1})$ is taken uniformly at random from the unit $(d-1)$-dimensional sphere in $\mathbb{R}^{d}$.

We conclude that:
\begin{align}
\begin{split}
\mathrm{Var}(K^{\mathrm{iid}}_{M})&-\mathrm{Var}(K^{\mathrm{ord}}_{M}) = 
\frac{4}{M^{2}(d+2)}\int_{\mathbb{R} \times ... \times \mathbb{R}} \sum_{1 \leq i < j \leq M} F_{x_{i}}(\mathbf{z})F_{x_{j}}(\mathbf{z}) \prod_{i=1}^{M} \mathcal{P}(x_{i}) \cdot dx_{1} \cdot ... \cdot dx_{M}\\
&= \frac{4}{M^{2}(d+2)} {M \choose 2}
\int_{\mathbb{R} \times \mathbb{R}} F_{x}(\mathbf{z})F_{y}(\mathbf{z})\mathcal{P}(x)\mathcal{P}(y)dx dy = (1-\frac{1}{M})\frac{2}{d+2}F^{2}(\mathbf{z}),
\end{split}
\end{align}
where $F(\mathbf{z}) = \mathbb{E}_{\mathbf{u} \sim \mathrm{Unif}(0,\mathcal{S}^{d-1})}\mathbb{E}_{x \sim \tilde{\Omega}(d)}\left[\tilde{g}(\mathbf{u}^{\top}\mathbf{z}, x)\right]$.
That completes the proof.
\end{proof}

\subsection{Proof of Theorem \ref{thm:tails}}
\label{app:tails}

\begin{proof}
To prove the first part of the theorem, we use the following Hoeffding's inequality:

\begin{lemma}[Hoeffding's Inequality]
Let $X_{1},...,X_{M}$ be $M$ independent random variables (not necessarily identically distributed) with zero mean. Assume furthermore that: $-a_{i} \leq X_{i} \leq b_{i}$ for $a_{i,}b_{i} \geq 0$ for $i=1,...,M$. Then the following is true for any $a>0$:
\begin{equation}
\mathbb{P}[|\sum_{i=1}^{M}X_{i}|>a] \leq 2 \cdot \exp \left(-\frac{a^{2}}{\sum_{i=1}^{N}(a_{i}+b_{i})^{2}} \right)    
\end{equation}
\end{lemma}
Note first that we have: 
\begin{align}
\begin{split}
\label{z-bound}
0 \leq Z = \exp\left(-\left\|\sqrt{-A}\omega-\frac{B}{2\sqrt{-A}}\mathbf{x}\right\|^{2}-\frac{B^{2}}{4A}\|\mathbf{x}\|^{2}+C\|\mathbf{x}\|^{2}\right) \cdot \\ 
\exp\left(-\left\|\sqrt{-A}\omega-\frac{B}{2\sqrt{-A}}\mathbf{y}\right\|^{2}-\frac{B^{2}}{4A}\|\mathbf{y}\|_{2}^{2}+C\|\mathbf{y}\|_{2}^{2}\right) \leq \exp \left(-\frac{\|\mathbf{x}\|^{2}+\|\mathbf{y}\|^{2}}{4A} \right),
\end{split}
\end{align}
where the last inequality follows from taking: $B=\sqrt{1-4A}$, $C=-1$. 

Denote: $\mathcal{M}(\mathbf{x},\mathbf{y})=\exp(-\frac{\|\mathbf{x}\|^{2}+\|\mathbf{y}\|^{2}}{4A})$. 
Define: $Y=Z-\mathbb{E}[Z]$. Note that: $\mathbb{E}[Y]=0$.
Furthermore, from Inequality \ref{z-bound}, we get:
$0 - K(\mathbf{x},\mathbf{y}) \leq Y \leq \mathcal{M}(\mathbf{x},\mathbf{y}) - K(\mathbf{x},\mathbf{y})$. Thus we have: $-a \leq Y \leq b$ for $a=K(\mathbf{x},\mathbf{y})$, $b=\mathcal{M}(\mathbf{x},\mathbf{y})-K(\mathbf{x},\mathbf{y})$. The following is true:
\begin{equation}
\mathbb{P}[|\widehat{K}^{\mathrm{iid}}_{M}(\mathbf{x},\mathbf{y})-K(\mathbf{x},\mathbf{y})| \geq \epsilon] = 
\mathbb{P} \left[\frac{Y_{1}+...+Y_{M}}{M} \geq \epsilon \right] = 
\mathbb{P}[|Y_{1}+...+Y_{M}| \geq M\epsilon],
\end{equation}
where $Y_{1}, \dots, Y_{M}$ are independent copies of $Y$. We complete the proof of the first part of the theorem by applying Hoeffding's Inequality for: $X_{i}=Y_{i}$, $a_{i}=a$, $b_{i}=b$ ($i=1,...,M$) and $a=M\epsilon$.

The second part of the theorem follows directly from the exact same method as applied in the proof of Theorem \ref{thm-orthoprfs} (e.g. we condition on the lengths of the sampled vectors $\omega_{i}$), combined again with the analysis from Theorem 5 in \cite{performer}, but this time for higher moments. Note that critically, Legendre Transform is well-defined since the corresponding random variables are bounded. The nonnegativity of the Legendre Transform for the inputs from statement of the theorem follows from the standard properties of the Legendre Transform for the inputs $x > \mathbb{E} X$, where $X$ is the corresponding random variable.
\end{proof}

\subsection{Proof of Theorem \ref{thm:uniform}}
\label{app:uniform}
\begin{proof}
The proof is similar to the proof of Claim 1 from \cite{rfs}.
Note that in the regular attention mechanism, queries and keys are renormalized by the multiplicative factor: $\frac{1}{d^{\frac{1}{4}}}$. Thus denote: $\mathbf{x}=\frac{\mathbf{q}}{d^{\frac{1}{4}}}$ and $\mathbf{y}=\frac{\mathbf{k}}{d^{\frac{1}{4}}}$. Note that: $\|\mathbf{x}\|,\|\mathbf{y}\| \leq \frac{R}{d^{\frac{1}{4}}}$. Consider vector $\mathbf{z} = [\mathbf{x}^{\top},\mathbf{y}^{\top}]^{\top} \in \mathbb{R}^{2d}$. Note that: $\|\mathbf{z}\|_{2} \leq \sqrt{2}\frac{R}{d^{\frac{1}{4}}}$. By the analogous analysis as in Claim 1, we cover the ball $B(0,\sqrt{2}\frac{R}{d^{\frac{1}{4}}}) \subseteq \mathbb{R}^{2d}$ with the $\epsilon$-net of at most $T=(\frac{4 \rho}{r})^{2d}$ balls of radius $r$ for $\rho=\sqrt{2}\frac{R}{d^{\frac{1}{4}}}$. If $L_{f}$ denotes the Lipschitz constant of $f$, the straightforward calculations lead to:
\begin{equation}
\mathbb{E}[L_{f}^{2}] \leq \max_{\mathbf{x},\mathbf{y}} \widehat{\mathcal{M}}^{2}(\mathbf{x},\mathbf{y})\max_{\mathbf{x},\mathbf{y}}\left(2\|\mathbf{x}\|^{2}+2\|\mathbf{y}\|^{2}+4\mathbb{E}[\|\omega\|_{2}^{2}]\right),    
\end{equation}
where $\widehat{\mathcal{M}}(\mathbf{x},\mathbf{y})=\exp(-\frac{\|\mathbf{x}\|_{2}^{2}+\|\mathbf{y}\|_{2}^{2}}{2})\mathcal{M}(\mathbf{x},\mathbf{y})$, $\mathcal{M}(\mathbf{x},\mathbf{y})$ is defined as in the proof above and $\omega \sim \mathcal{N}(\mathbf{0}_d, \mathbf{I}_{d})$ (the extra multiplicative term next before $\mathcal{M}(\mathbf{x},\mathbf{y})$ is needed since now we work with the softmax-kernel which is the rescaled variant of the Gaussian kernel, see: discussion in the paper). Thus we have: $\mathbb{E}[L_{f}^{2}] \leq \gamma^{2}$, where: 
$\gamma=2\sqrt{\exp(-\frac{3R^{2}}{A\sqrt{d}})(\frac{R^{2}}{\sqrt{d}}+d^{2})}$. Using Theorem \ref{thm:tails}, we also notice that we can get analogous inequality as Inequality (6) from the proof of Claim 1 in \cite{rfs}, but for:
$D=4M\max_{\mathbf{x},\mathbf{y}}\exp(\frac{3(\|\mathbf{x}\|_{2}^{2}+\|\mathbf{y}\|_{2}^{2})}{2A})=4M\exp(\frac{3R^{2}}{A\sqrt{d}})$. Thus substituting: (a) $\sigma_{p}$ with $\gamma$, (b) $D$ with $4M\exp(\frac{3R^{2}}{A\sqrt{d}})$, (c) $d$ with $2d$ and (d) $\mathrm{diam}(\mathcal{M})$ with $\rho$ in the statement of Claim 1, we obtain Theorem \ref{thm:uniform}. 

\end{proof}

\subsection{Additional experimental details}
\label{app:exp}

We use NumPy \cite{numpy} and the free version of Google Colaboratory for running the first two experiments. For the Transformer experiments, we use a TPU cluster and JAX \cite{jax} implementation.

\subsubsection{Comparing variance of different RFs} \label{app:varexp}

We use Brent method \cite{brent} with $100$ iterations for minimization of $p$ in GeomRF(+) and two L-BFGS-B \cite{lbfgsb} routines of $50$ iterations to minimize $A$ in GERF for $s = -1$ and $+1$ respectively. We reuse these configurations in the non-parametric classification experiment.

We sample pairs of sets $\{ \*x^{(i)} \}_{1 \leq i \leq L}$, $\{ \*y^{(j)} \}_{1 \leq j \leq L}$, where $L = 1024$, $5$ times. On each pair of sets, we compute the variance of approximating $K (\*x^{(i)}, \*y^{(j)})$ for all pairs of $\*x^{(i)}$ and $\*y^{(j)}$. Also, on each pair of sets of $\{ \*x^{(i)} \}_{1 \leq i \leq L}$, $\{ \*y^{(j)} \}_{1 \leq j \leq L}$, we compute statistics (\ref{eq:gestat},\ref{eq:poisstat},\ref{eq:geomstat}) and then use them to optimize parameters of the corresponding method. The means and standard deviations are reported for averaging over all pairs of $\*x^{(i)}$ and $\*y^{(j)}$, over all $5$ samples.

For a fair comparison, for real-valued RF mechanisms, we compute the variance assuming that $M = 2$ (the variance is divided by $2$), since complex RF mechanisms effectively use real and imaginary part of the number.

\subsubsection{Non-parametric classification} \label{app:classexp}

We randomly split the raw dataset into $90\%$ which is used for training, $5\%$ for tuning $\sigma$ and $5\%$ for testing. These splits are fixed for all compared methods. $\sigma$ is tuned on a log-uniform grid of $10$ values from $10^{-2}$ to $10^2$. For each $\sigma$ and each method, we average accuracy for $50$ seeds used to draw RFs both during validation and testing (for the best $\sigma$ only). As for the previous experiment, we use $M = 128$ for real-valued RFs and $M = 64$ for complex-valued for a fair comparison. We use orthogonal $\omega$'s for all GERF-descendant methods.  Table \ref{tab:ucistd} reports standard deviations of the test accuracies reported in the main text.

We use $\epsilon = 10^{-8}$ when making input features positive in PoisRF+ and GeomRF+. $\*c$ is inferred from the train set, and we clamp validation/test input features to be at least $\epsilon$ to guarantee that they are positive without leaking test data into $\*c$.

\begin{table}[t]
\small
\centering
\caption{Non-parametric classification, standard deviations.}
\label{tab:ucistd}
\begin{tabular}{@{}l|cc|cccccc@{}}
\toprule
Dataset & TrigRF & PosRF & GERF & PoisRF & GeomRF & OPRF & PoisRF+ & GeomRF+ \\
\midrule
\texttt{abalone} & $< 0.05$ & $2.1$ & $1.9$ & $1.8$ & $1.3$ & $1.7$ & $2.9$ & $2.9$ \\
\texttt{banknote} & $< 0.05$ & $3.7$ & $4.3$ & $2.1$ & $3.0$ & $3.4$ & $5.9$ & $7.7$ \\
\texttt{car} & $< 0.05$ & $3.0$ & $2.5$ & $0.0$ & $< 0.05$ & $3.0$ & $< 0.05$ & $1.5$ \\
\texttt{yeast} & $< 0.05$ & $3.2$ & $5.0$ & $6.0$ & $3.4$ & $4.9$ & $< 0.05$ & $2.4$ \\
\texttt{cmc} & $< 0.05$ & $4.0$ & $3.9$ & $4.3$ & $3.4$ & $3.8$ & $5.3$ & $5.2$ \\
\texttt{nursery} & $< 0.05$ & $6.3$ & $3.2$ & $7.2$ & $7.3$ & $6.3$ & $5.6$ & $8.2$ \\
\texttt{wifi} & $< 0.05$ & $6.2$ & $4.1$ & $2.8$ & $2.0$ & $4.1$ & $13.1$ & $9.8$ \\
\texttt{chess} & $< 0.05$ & $1.3$ & $1.2$ & $1.8$ & $2.0$ & $1.2$ & $1.7$ & $1.9$ \\
\bottomrule
\end{tabular}
\end{table}

\subsubsection{Text \label{sec:appedinx_text}} 

We pretrained on two publicly available datasets (see: Table \ref{tab:mlm_data}). Following the original BERT training, we mask $15\%$ of tokens in these two datasets, and train to predict the mask. We used the exact same hyperparameter-setup for all the baselines (FAVOR+ \cite{performer}, ELU \cite{pmlr-v119-katharopoulos20a}, ReLU \cite{performer}) and FAVOR++. The hyperparameters for pretraining are shown in Table \ref{tab:app_mlm_param}. 
We finetuned on GLUE task, warm-starting with the weights of the pretrained model. The setup is analogous to the one from the original BERT paper.

\begin{table}[h]
\small
\centering
\caption{Hyperparameters for the base models for pre-training for the baselines (FAVOR+ \cite{performer}, ELU \cite{pmlr-v119-katharopoulos20a} and ReLU \cite{performer}) and FAVOR++.}
\label{tab:app_mlm_param}
\begin{tabular}{@{}l c r c r @{}}
\toprule
Parameter & & Value  \\
\midrule
 $\#$ of heads & & $12$ \\
 $\#$ of hidden layers & & $12$  \\
 Hidden layer size & & $768$  \\
 $\#$ of tokens & & $512$ \\
 Batch size & & $256$  \\
 M & & $256$ \\
 Pretrain Steps & & $1M$ \\
 Loss & & MLM  \\
 Activation layer & & gelu \\ 
 Dropout prob & & $0.1$  \\
 Attention dropout prob & & $0.1$ \\
 Optimizer & & Adam \\
 Learning rate & & $10^{-4}$  \\
 Compute resources & & $8 \times 8$ TPUv3 \\
\bottomrule
\end{tabular}
\end{table}

\begin{table}[h]\vspace{-3mm}
    \centering
    \small
    \centering
    \caption{Dataset used for pre training.}
    \label{tab:mlm_data}
    \begin{tabular}{@{}lrr@{}}
    \toprule
    Dataset & $\#$ tokens & Avg. doc len. \\
    \midrule
        Books \cite{zhu2015aligning} & $1.0$B & $37$K \\
        Wikipedia &  $3.1$B & $592$ \\
    \bottomrule
    \end{tabular}
    \vspace{1mm}
  \vspace{-3mm}
\end{table}

\subsubsection{Speech}

Our applied Conformer-Transducer models consisted of $l=17$ conformer layers. Each attention layer used $h=4$ heads. The embedding dimensionality was $p=256$. Dimensions were split equally among heads, leading to $d_{QK}=64$ dimensions per query/key. Input sequences were of length $L \sim 500$. We applied padding mechanism for all tested variants. The model provides transcribed speech (see also: Table: \ref{tab:app_speech_param}).

\begin{table}[h]
\small
\centering
\caption{Hyperparameters for trained Speech models.}
\label{tab:app_speech_param}
\begin{tabular}{@{}l c r c r @{}}
\toprule
Parameter & & Value  \\
\midrule
 $\#$ of heads & & $4$ \\
 $\#$ of hidden layers & & $17$  \\
 Hidden layer size & & $256$  \\
 $\#$ of tokens & & $512$ \\
 Batch size & & $256$  \\
 Activation layer & & gelu \\ 
 Dropout prob & & $0.1$  \\
 Optimizer & & Adam \\
 Learning rate & & $10^{-4}$  \\
 Compute resources & & $8 \times 8$ TPUv3 \\
\bottomrule
\end{tabular}
\end{table}

\subsubsection{Vision \label{sec:appendix_vision}} 
The vision experiments follow Section 4 in the MAE paper, where we use a ViT-Large (Table: ~\ref{tab:vitlarge}) and the same setup for training from scratch (Table: ~\ref{tab:vitscratch}) and fine-tuning (Table: ~\ref{tab:vitfinetune}) as for the MAE baseline trained with regular softmax attention (Table: ~\ref{tab:vitpretrain}). Note that the fine-tuning setup has a shorter schedule which tests the adaptability of low-rank attention variants to the regular softmax attention.     

The ablations over sequence lengths are conducted by training from scratch and use ViT-tiny model (Table: ~\ref{tab:vittiny}). Different sequence lengths are derived by adjusting the input size and the patch size which results in different number of patches (Table: ~\ref{tab:vitpatch}). Different patch sizes require different sizes of projection layers before converting to tokens with latent representations of the same dimesionality. 

\begin{table}[h]
\small
\centering
\caption{Hyperparameters for Vision pre-training setting.}
\label{tab:vitpretrain}
\begin{tabular}{@{}l c r c r @{}}
\toprule
Parameter & & Value  \\
\midrule
 Batch size & & $4096$  \\
 Optimizer & & AdamW \\
 Base Learning rate & & $1.5e^{-4}$  \\
 Weight decay & & 0.05 \\
 Optimizer momentum & & $\beta_1, \beta_2$ = 0.9, 0.95  \\
 Learning rate schedule & & cosine decay \\
 Warm up epochs & & 40 \\
 Augmentation & & RandomResizedCrop \\
 Compute resources & & $8 \times 8$ TPUv3 \\
\bottomrule
\end{tabular}
\end{table}

\begin{table}[h]
\small
\centering
\caption{Hyperparameters for Vision End-to-End fine-tuning setting.}
\label{tab:vitfinetune}
\begin{tabular}{@{}l c r c r @{}}
\toprule
Parameter & & Value  \\
\midrule
 Batch size & & $1024$  \\
 Optimizer & & AdamW \\
 Base Learning rate & & $1e^{-3}$  \\
 Layer-wise lr decay && 0.75 \\
 Weight decay & & 0.05 \\
 Optimizer momentum & & $\beta_1, \beta_2$ = 0.9, 0.999  \\
 Learning rate schedule & & cosine decay \\
 Warm up epochs & & 5 \\
 Training epochs & & 50 \\
 Augmentation & & RandomAug (9, 0.5) \\
 Label smoothing && 0.1 \\
 Mixup && 0.8 \\
 CutMix && 1.0 \\
 Droppath && 0.1 \\
 Compute resources & & $8 \times 8$ TPUv3 \\
\bottomrule
\end{tabular}
\end{table}

\begin{table}[h]
\small
\centering
\caption{Hyperparameters for Vision - training from scratch setting.}
\label{tab:vitscratch}
\begin{tabular}{@{}l c r c r @{}}
\toprule
Parameter & & Value  \\
\midrule
 Batch size & & $4096$  \\
 Optimizer & & AdamW \\
 Base Learning rate & & $1e^{-4}$  \\
 Layer-wise lr decay && 0.75 \\
 Weight decay & & 0.3 \\
 Optimizer momentum & & $\beta_1, \beta_2$ = 0.9, 0.999  \\
 Learning rate schedule & & cosine decay \\
 Warm up epochs & & 20 \\
 Training epochs & & 200 \\
  Augmentation & & RandomAug (9, 0.5) \\
 Label smoothing && 0.1 \\
 Mixup && 0.8 \\
 CutMix && 1.0 \\
 Droppath && 0.2 \\
 Exp moving avg  && 0.9999 \\
 Compute resources & & $8 \times 8$ TPUv3 \\
\bottomrule
\end{tabular}
\end{table}

\begin{table}[h]
\small
\centering
\caption{Hyperparameters for Vision model - ViT Large.}
\label{tab:vitlarge}
\begin{tabular}{@{}l c r c r @{}}
\toprule
Parameter & & Value  \\
\midrule
 $\#$ of heads && 16 \\
 $\#$ of layers && 24 \\
 Hidden layer size && 1024 \\
\bottomrule
\end{tabular}
\end{table}

\begin{table}[h]
\small
\centering
\caption{Hyperparameters for Vision model - ViT tiny.}
\label{tab:vittiny}
\begin{tabular}{@{}l c r c r @{}}
\toprule
Parameter & & Value  \\
\midrule
 $\#$ of heads && 3 \\
 $\#$ of layers && 12 \\
 Hidden layer size && 192 \\
\bottomrule
\end{tabular}
\end{table}

\begin{table}[h]
\small
\centering
\caption{ViT sequence length ($\#$ patches) and image input mapping.}
\label{tab:vitpatch}
\begin{tabular}{@{}l c r c r @{}}
\toprule
Patches & & Image input size  \\
\midrule
8x8 && 224 \\
16x16 && 224 \\
32x32 && 224 \\
40x40 && 240 \\
44x44 && 220 \\
\bottomrule
\end{tabular}
\end{table}

\end{document}